\documentclass[journal]{IEEEtran}
\usepackage{multirow} 
\usepackage{amsmath,amsfonts}
\usepackage{algorithmic}
\usepackage{algorithm}
\usepackage{array}
\usepackage{booktabs}
\usepackage[caption=false,font=normalsize,labelfont=sf,textfont=sf]{subfig}
\usepackage{textcomp}
\usepackage{stfloats}
\usepackage{url}
\usepackage{verbatim}
\usepackage{graphicx}
\usepackage{cite}
\usepackage{amsmath}
\usepackage[table,xcdraw]{xcolor}
\usepackage{amssymb}
\usepackage{xcolor}
\definecolor{rblue}{rgb}{0,0.5,1}
\definecolor{hollywoodcerise}{rgb}{0.96, 0.0, 0.63}
\definecolor{lasallegreen}{rgb}{0.03, 0.47, 0.19}
\definecolor{hanpurple}{rgb}{0.32, 0.09, 0.98}
\definecolor{green(pigment)}{rgb}{0.0, 0.65, 0.31}
\usepackage[pagebackref=false,breaklinks=true,colorlinks,bookmarks=false]{hyperref}
\hypersetup{colorlinks,linkcolor={red},citecolor={hanpurple},urlcolor={magenta}}  
\DeclareUnicodeCharacter{2061}{}

\begin{document}

\title{DTCLMapper: Dual Temporal Consistent Learning for Vectorized HD Map Construction}

\author{Siyu Li$^{1}$, Jiacheng Lin$^{2}$, Hao Shi$^{3}$, Jiaming Zhang$^{4}$, Song Wang$^{5}$, You Yao$^{6}$, Zhiyong Li$^{1,2}$, and Kailun Yang$^{1}$
\thanks{This work was supported in part by the National Natural Science Foundation of China (No. U21A20518, No. 61976086, and No. 62473139) and in part by Hangzhou SurImage Technology Company Ltd. \textit{(Corresponding authors: Zhiyong Li and Kailun Yang.)}}
\thanks{$^{1}$S. Li, Z. Li, and K. Yang are with the School of Robotics and the National Engineering Research Center of Robot Visual Perception and Control Technology, Hunan University, Changsha 410082, China (email: zhiyong.li@hnu.edu.cn; kailun.yang@hnu.edu.cn).}%
\thanks{$^{2}$J. Lin and Z. Li are with the College of Computer Science and Electronic Engineering, Hunan University, Changsha 410082, China.}%
\thanks{$^{3}$H. Shi is with the State Key Laboratory of Extreme Photonics and Instrumentation and the National Engineering Research Center of Optical Instrumentation, Zhejiang University, Hangzhou 310027, China.}%
\thanks{$^{4}$J. Zhang is with the Institute for Anthropomatics and Robotics, Karlsruhe Institute of Technology, Karlsruhe 76131, Germany.}%
\thanks{$^{5}$S. Wang is with the College of Computer Science, Zhejiang University, Hangzhou 310027, China.}%
\thanks{$^{6}$Y. Yao is with the USC Viterbi School of Engineering, the University of Southern California, Los Angeles 90089, California, United States.}%
}

\markboth{IEEE Transactions on Intelligent Transportation Systems, August~2024}%
{Li \MakeLowercase{\textit{et al.}}: Dual Temporal Consistent Learning.}

\maketitle

\begin{abstract}
Temporal information plays a pivotal role in Bird's-Eye-View (BEV) driving scene understanding, which can alleviate the visual information sparsity. However, the indiscriminate temporal fusion method will cause the barrier of feature redundancy when constructing vectorized High-Definition (HD) maps. In this paper, we revisit the temporal fusion of vectorized HD maps, focusing on temporal instance consistency and temporal map consistency learning. To improve the representation of instances in single-frame maps, we introduce a novel method, DTCLMapper. This approach uses a dual-stream temporal consistency learning module that combines instance embedding with geometry maps. In the instance embedding component, our approach integrates temporal Instance Consistency Learning (ICL), ensuring consistency from vector points and instance features aggregated from points. A vectorized points pre-selection module is employed to enhance the regression efficiency of vector points from each instance. Then aggregated instance features obtained from the vectorized points preselection module are grounded in contrastive learning to realize temporal consistency, where positive and negative samples are selected based on position and semantic information. The geometry mapping component introduces Map Consistency Learning (MCL) designed with self-supervised learning. The MCL enhances the generalization capability of our consistent learning approach by concentrating on the global location and distribution constraints of the instances. Extensive experiments on well-recognized benchmarks indicate that the proposed DTCLMapper achieves state-of-the-art performance in vectorized mapping tasks, reaching $61.9\%$ and $65.1\%$ mAP scores on the nuScenes and Argoverse datasets, respectively. The source code is available at \url{https://github.com/lynn-yu/DTCLMapper}.

\end{abstract}


\section{Introduction}
\IEEEPARstart{H}{igh-Definition} maps (HD maps) represent one of the cornerstones of autonomous driving tasks, which provides prior knowledge about static obstacles. The HD map is constructed in Bird's-Eye-View (BEV) space translated from perspective image views~\cite{bevsurvey}, similar to top-view semantic understanding~\cite{mass} for LiDAR data and camera-LiDAR fusion~\cite{camlid}.
Semantic maps generally use grids to depict roads on a plane, whereas vectorized maps encapsulate road structures with geometric points and lines. 
Although both types serve distinct purposes, vectorized maps offer a significant advantage--they require less storage while achieving higher accuracy, a particularly valuable trait for online perception tasks in autonomous driving~\cite{survey}. 

\begin{figure}[tb]
      \centering
      \includegraphics[scale=0.52]{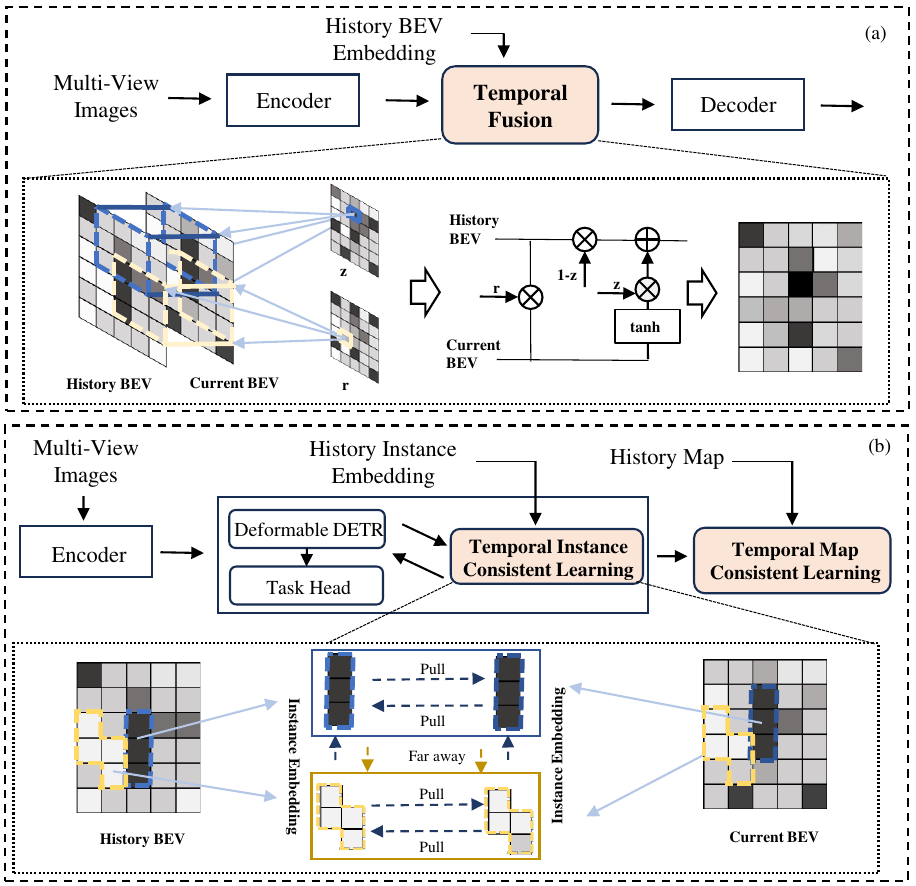}
      \caption{Difference between current temporal fusion and the proposed DTCLMapper. (a) depicts an example of the current BEV temporal fusion study~\cite{beverse} where features are enhanced through temporal fusion for intermediate BEV feature layers. (b) depicts the proposed DTCLMapper for BEV HD map construction. It can enrich road information from historical instance features using a consistency learning approach.
      }
      \label{Fig.1}
      \vspace{-2.0em}
\end{figure}

As a pioneering work, HDMapNet~\cite{hdmapnet} employed a simple MultiLayer Perceptron (MLP) as the view transformer module and used a basic ResNet structure as the semantic and instance segmentation head.
Unlike this work, where vectorized HD maps are built through postprocessing, MapTR~\cite{maptr} has developed an end-to-end model suitable for vectorized HD map construction. 
Despite their high performances, a known and essential problem in visual HD map research lies in the sparsity of information. 
Specifically, it is difficult to present the rich spatial information of the 3D world (or BEV space) through the 2D perspective view where the object occlusion phenomenon exists in the complex driving surroundings. 

\begin{figure}[tb]
      \centering
      \includegraphics[scale=0.42]{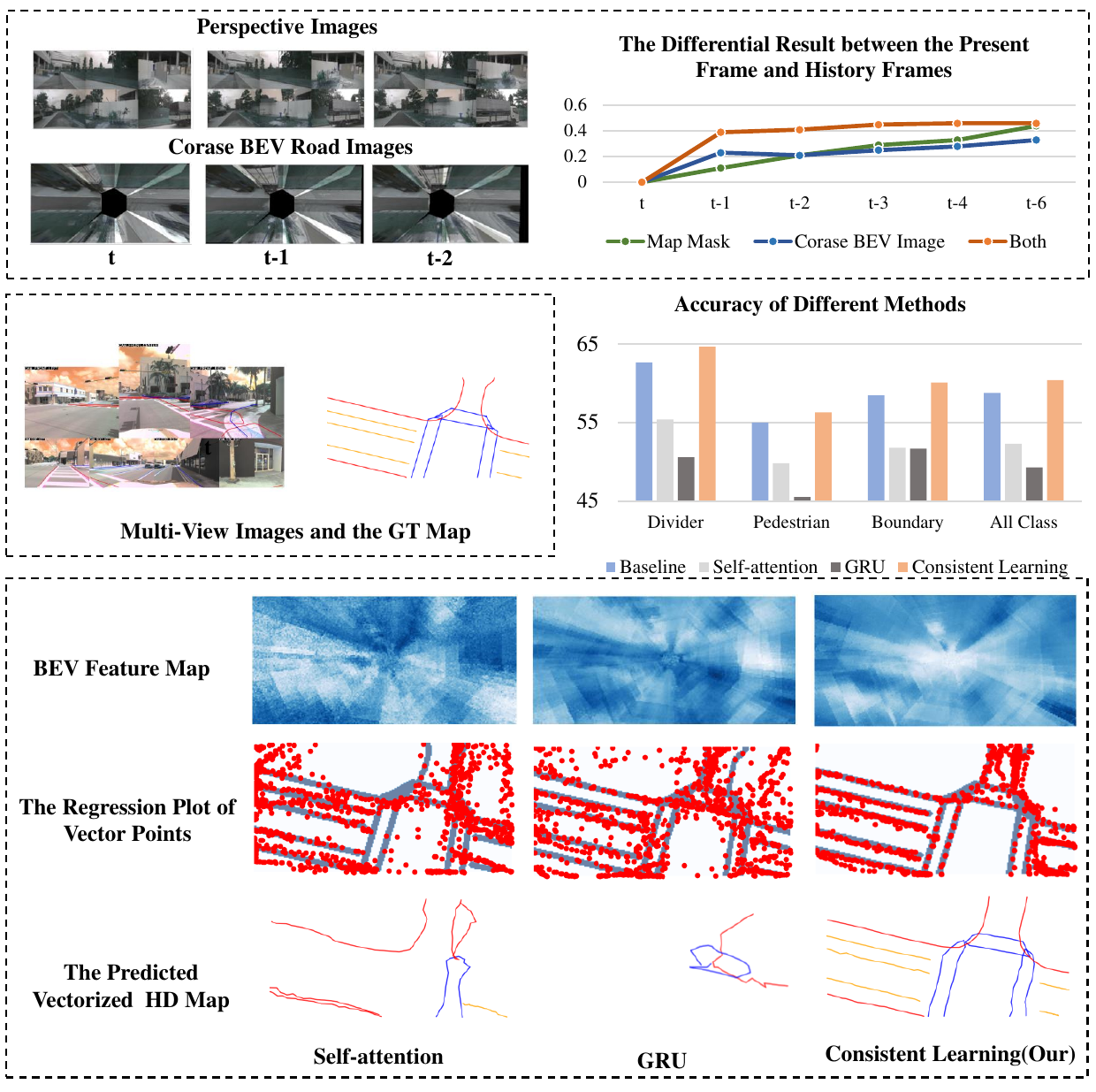}
      \vspace {-1.0em}
      \caption{Analysis of temporal overlapping and results of different temporal methods. Successive perspective images and coarse BEV road images through Inverse Perspective Mapping (IPM) are presented on the top left, whereas on the right are the differential results of successive frames. It can be observed that the BEV image in the short temporal range has a small difference. The middle and bottom sub-figures show comparison results of different temporal fusion methods. Intuitively, from the perspectives of point regression efficiency, map generation quality, and accuracy of three levels, the proposed temporal learning has excellent performance for HD map construction.
      }
      \label{Fig.intro}
      \vspace {-2.0em}
\end{figure}

Recent studies~\cite{BEVFormer,beverse} have shown that online temporal BEV fusion is an effective solution to tackle this challenge in the field of dynamic object detection. 
As shown in Fig.~\ref{Fig.1}, in these existing works, historical BEV information is aggregated to this moment through self-attention~\cite{attention}, or Gate Recurrent Unit (GRU)~\cite{gru}. 
This is defined as a ``temporal fusion strategy''. 
However, as shown in the middle and bottom of Fig.~\ref{Fig.intro}, these temporal fusion strategies for BEV learning from scratch have little impact on improving the quality of vectorized HD maps. 
Specifically, the characteristics of the integrated BEV feature close to the map instance are not discernible, causing the vector point to struggle to return to its precise position, thereby impacting the detection accuracy negatively. 
To delve deeper into the reasons, we analyze the overlapping rate of static instances in consecutive frames, as shown in the top of Fig.~\ref{Fig.intro}, where the differences in short consecutive frames are relatively small. 
During the process of BEV learning from scratch, facing similar features, direct merging strategies make it difficult to distinguish reliable characteristics, leading to feature redundancy issues.
In other words, the relationships and differences among individual objects, are still not adequately explored in terms of their dissimilarity and similarity.

To address this challenge, we move away from prior indiscriminate temporal fusion strategies and examine how the online enhancement of sparse instance features can be achieved through the consistency of instance targets across consecutive frames. 
This approach focuses on preserving unique characteristics while reinforcing connections between instances in successive frames. 
First, we present DTCLMapper, which features two consistent learning components: Instance Consistent Learning (ICL) and Map Consistent Learning (MCL). These components have a progressive and complementary relationship. 
On the one hand, ICL is composed of a Vector Point PreSelection Module (VPPSM) and Aggregated Instance Feature Consistent Learning (AIFCL), where VPPSM provides accurate instance features for consistent learning in instances and AIFCL promotes the consistent enhancement of sparse instances. 
On the other hand, the MCL component enhances the model by utilizing consistent geometric positions among instances. 
It uses a grid map rasterized from a vectorized map to enforce consistency, with map occupancy loss as the measurement mechanism, ensuring spatial relationships are maintained and improving model consistency and generalization. 
Extensive experiments on public nuScenes~\cite{nus} and Argoverse~\cite{argoverse} datasets, demonstrate that our method has greatly enhanced HD map construction, setting state-of-the-art performances of $61.9\%$ and $65.1\%$ in mAP, respectively.

The main contributions delivered in this work are summarized as follows:
\begin{itemize}
\item We design a dual-temporal consistent module for constructing vectorized HD maps called DTCLMapper, which addresses the sparsity of visual information in HD map construction by utilizing temporal consistency. To the best of our knowledge, this is the first study to employ a temporal-consistency-based strategy to tackle this challenge in the context of HD map construction.
\item A novel Instance Consistent Learning (ICL) composed of a VPPSM and AIFCL is proposed to improve the accuracy of instances underlying consistent learning. Specifically, through the positive samples with high similarity and the negative samples with different labels, the consistency of complementary reinforcement learning of the same instance is promoted.
\item We propose a novel Map Consistent Learning (MCL) to enforce consistency across global geometry and states by evaluating the occupancy status of trusted vector instances. MCL helps maintain a consistent spatial arrangement of instances, ensuring that the global map accurately reflects the relationships among various components in terms of their geometry and occupancy.
\item Extensive experiments on well-recognized datasets demonstrate the superiority of our method and the effectiveness of the key components.
\end{itemize}

\section{Related Work}
\label{related}

\textbf{View Transformation in BEV Understanding:} 
In the HD map construction research field, the view transformation between perspective and BEV view is an important module. There is a wealth of excellent  studies~\cite{camera,vit} work in the field of perspective view for driving scene understanding. And a variety of pipelines translating perspective views into BEV views have been proposed in the community, ranging from IPM~\cite{ipm}, depth estimation~\cite{LSS,bevdepth}, Variational Auto-Encoder (VAE)~\cite{ved}, and learning-based parameter optimization methods~\cite{vpn,cvt,trans2map,BEVFormer,ibevseg}. 
LSS~\cite{LSS} estimated the depth of each pixel through the appearance relation of adjacent objects. 
Then the pixel features were projected into the 3D space by the intrinsic and extrinsic parameters. Naturally, the BEV feature can be obtained simply in the 3D space. 
Furthermore, considering the sparsity of depth estimation, BEVDepth~\cite{bevdepth} 
demonstrated that supervising depth estimation with depth values from laser point clouds effectively enhances the expression of BEV features, surpassing even transformer-based frameworks.
VPN~\cite{vpn} investigated the potential of using MLP to realize view transformation, which implicitly included depth estimation and camera parameters in the network learning.
BEVFormer~\cite{BEVFormer} has proposed a classic of a different kind of work, which projected 3D points directly into 2D-pixel spaces to learn BEV features. Simultaneously, this work has introduced the temporal fusion operation. 
Specifically, the attention operation was used to learn the historical features initially, and then the current-moment features were learned. 
Recently, the improvement of BEVFormer in terms of temporal fusion was presented in~\cite{BEVFormerv2}. Instead of merging temporal before current learning, it merged temporal BEV features directly by concatenating operations on channels. Similarly, BEVerse~\cite{beverse} designed temporal fusion for BEV features, except that GRU operations are used. All these temporal works perform indiscriminate learning at the BEV feature level. Although different features can be supplemented, similar features are prone to feature redundancy. 
In particular, in the HD map research, it does not bring a positive impact. Unlike existing works, we propose consistent learning for temporal instances. 
The idea originates from the consistency of map instances, with the aim of complementing the diverse characteristics of map instances for enhancing HD map construction.

\textbf{Online HD Map Learning:} 
In an HD map, road dividers, pedestrian crossing, and road boundaries are basic semantic information. In past studies, HD map was generated in an offline manner, which is time-consuming and labor-intensive work~\cite{slamreview}. Witnessed a rapid growth of BEV understanding, HDMapNet~\cite{hdmapnet} was the first to construct an online BEV HD map where MLP was leveraged to be the view transformation. It introduced the semantic HD map characterized by a grid and the vectorized HD map comprised of vector points and lines. 
Then, a variety of research works~\cite{bevsegformer,bimapper,mvmap,neruralmap,pmapnet}, have made extensive efforts to enhance the performance of semantic HD map construction. 
The aforementioned online map research objects are divided into two categories, a semantic map and a vectorized map. Different from a semantic map, a vectorized map is composed of vector points and lines that have fewer memory requirements. 
Moreover, it can provide basic road information for tasks such as trajectory prediction~\cite{tra-pre,w1,w2} and navigation~\cite{image} of intelligent vehicles.
Unlike the work~\cite{hdmapnet}, where the vectorized HD map was constructed through post-processing, the proposed method employs an end-to-end vectorized map generation method.

\textbf{Vectorized HD Map:} 
There have been significant advancements in studying an end-to-end method. 
These works are classified into two categories according to the expression of vector lines. 
One focuses on the Bezier curve that is composed of a start point, a stop point, and a control point~\cite{stsu,bezier-1,bezier-2}.
STSU~\cite{stsu}, as the classic work, depicted the directed graph of the centerline line of the lane through the Bezier curve. 
It followed the transformer-based object detector method with two learning queries, namely, centrelines and instance objects. 
Then, the work of~\cite{bezier-1} designed a novel form of the topology road map based on STSU. 
To model a concise and elegant Bezier curve, BeMapNet~\cite{bezier-2} proposed piecewise Bezier curves. 
On the other hand, VectorMapNet~\cite{vecmapnet} reckoned that polylines are more suitable as the basic element of a vector map. 
In VectorMapNet, the points of polylines are generated through an autoregressive model based on a transformer decoder. 
In addition, they proposed a novel paradigm about key point representations of map elements, which is the cornerstone of the following works.

MapTR~\cite{maptr} leveraged Deformable DETR~\cite{ddetr} as a decoder and can identify points with different orderings. 
It greatly advanced the real-time capabilities of map generation.
Recently, they optimized the form of query in training, which applied a mixed strategy for each query to enhance mapping accuracy~\cite{maptrv2}.
There have been many studies that have refined this work over time~\cite{insightmapper,himap,mgmap}. 
InsightMapper~\cite{insightmapper} modified the query mechanism that is translated into a hybrid generation mechanism.
PivotNet~\cite{pivotnet} designed a dynamic matching to obtain key points for each vector line.
InstaGraM~\cite{instmap} utilized graph construction to learn points and lines in a map. As sparse points lack precision in subtle places, MapVR~\cite{vect2raster} designed a rasterization-based metric to improve the performance of a vectorized map.
In addition, 3D lane detection~\cite{3dlanenet,bevlanedet,transbevlane,hdmapgen} can obtain a more accurate road structure.
Recently, some scholars have explored the fusion and tracking of large-scale temporal memory buffers. 
StreamMapNet~\cite{streammapnet} introduced a streaming strategy, which is different from our stacking strategy. It blends all the visible features of the past at the feature level and joins high-confidence instances from historical temporal data.
MapTracker~\cite{maptracker} supervised the association of map instances and fused traceable objects in historical frames from BEV feature warming, instance vector warming, and joint training perspectives. 
In contrast to the two methods, we focus on short-term BEV features learned from scratch. Our work introduces a unique temporal processing approach during the training phase, enhancing the features of element instances through temporal consistent learning and bypassing complex temporal parameter calculations in the testing phase, which is more flexible and suitable for real-world online HD mapping.

\textbf{Contrastive Learning:} 
Contrastive learning is a type of unsupervised learning that significantly advances unsupervised tasks~\cite{pcont}. 
SimCLR~\cite{SimCLR} designed a data enhancement idea and a nonlinear contrast feature learning module, improving the quality of learning representations.
However, due to SimCLR requiring a large set of negative samples in contrast learning, MOCO~\cite{moco} used momentum encoders to reduce the number of negative samples. 
MOCOv2~\cite{mocov2} followed a nonlinear contrast feature learning module in SimCLR~\cite{SimCLR} to further upgrade the unsupervised learning accuracy. 
At the same time, contrastive learning also shines brightly in the study of supervised learning~\cite{end,cls,idol,ctvis}. Compared with MOCO where samples were obtained by data augmentation, these work selected samples under supervision. 
In these studies, contrastive learning for temporal information has also been extensively explored, especially in video instance segmentation. 
IDOL~\cite{idol} proposed an online framework based on contrastive learning, which improved the discriminative ability of instances by associating information between temporal series. 
Based on it, CTVIS~\cite{ctvis} raised an effective training strategy to learn consistency for challenging targets in consecutive frames, such as occlusion and deformation.

Interestingly, we have observed that the temporal learning of video instance segmentation is similar to HD map construction. Coupled on reference frames, these video segmentation models~\cite{idol,ctvis} use contrastive learning to find the correlation object in the target frame, achieving online semantic segmentation. 
In constructing HD maps, the temporal analysis also aimed for scenarios with highly correlated instances, similar to the associative learning of videos. However, the temporal cues are different between them. The former focuses on objects that are dynamic with specific instance information, whereas this work addresses static instance objects based on uncertain BEV features. Motivated by these similarities and differences, we explore the potential usability of contrastive learning in HD map construction. We first enhance the accuracy of vectorized instance mapping and then utilize the consistency of static objects to establish a temporal contrastive model.

\begin{figure*}[tb]
      \centering
      \includegraphics[width=\linewidth]{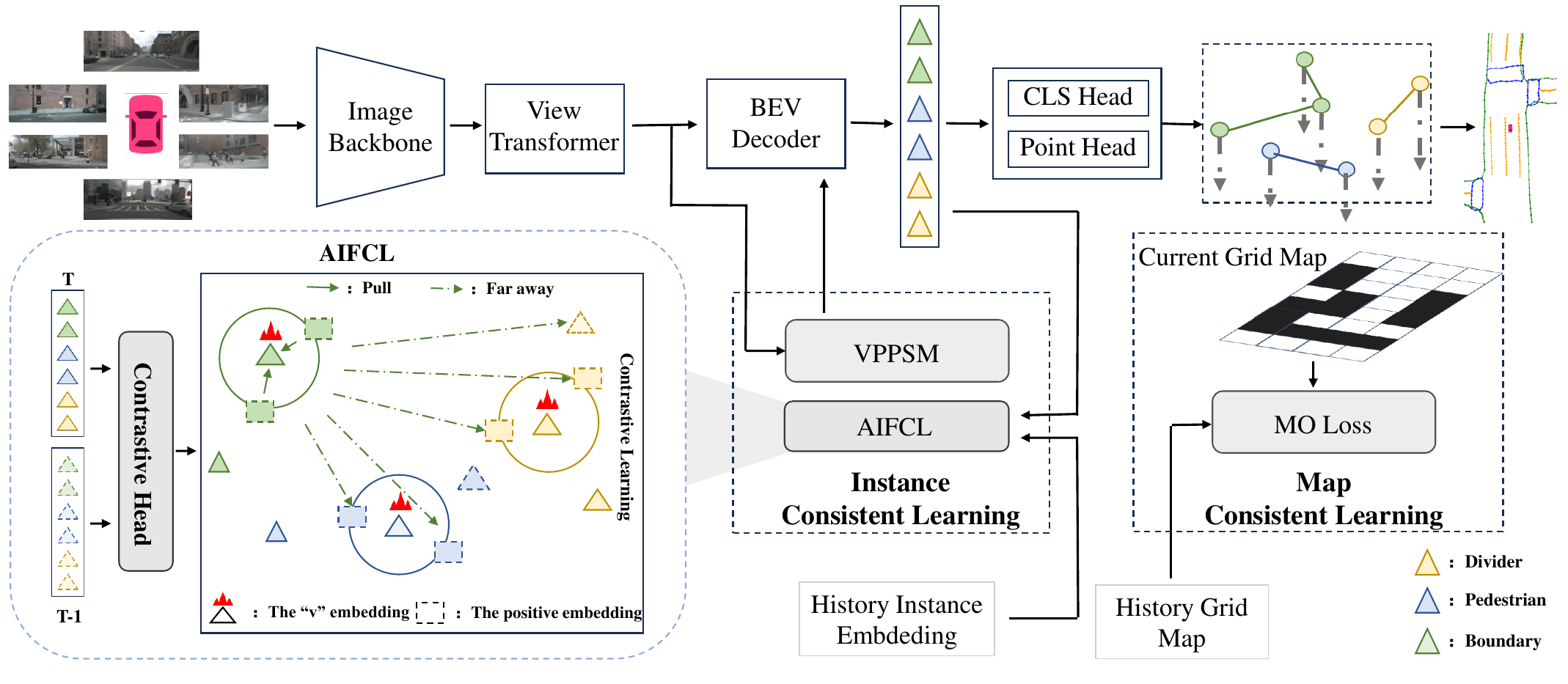}
      \caption{Overview of the proposed DTCLMapper architecture. It consists of the multi-view image backbone, view transformer, BEV decoder, and multi-heads. To alleviate the sparsity of visual information, a dual temporal consistent learning module is introduced, namely, Instance Consistent Learning (ICL) and Map Consistent Learning (MCL). ICL is composed of a Vector Point PreSelection Module (VPPSM) and Aggregated Instance Feature Consistent Learning (AIFCL). MCL proposes a Map Occupancy Loss (MO Loss) based on a grid map.}
      \label{Fig.2}
      \vspace{-1.7em}
\end{figure*}
\section{Method}
This section first introduces the framework of DTCLMapper (Sec.~\ref{sec:framwork}). 
We then propose the temporal Instance Consistent Learning (ICL) (Sec.~\ref{sec:il}), including the introduction of a Vector Point PreSelection Module (VPPSM) that can enhance the accuracy of instances. 
Additionally, the Contrastive Learning of Aggregated Instance Features (AIFCL) is applied to enrich the expression of instances. 
Finally, the details of Map Consistent Learning (MCL) are introduced (Sec.~\ref{sec:ml}). 

\begin{figure*}[tb]
      \centering
      \includegraphics[width=\linewidth]{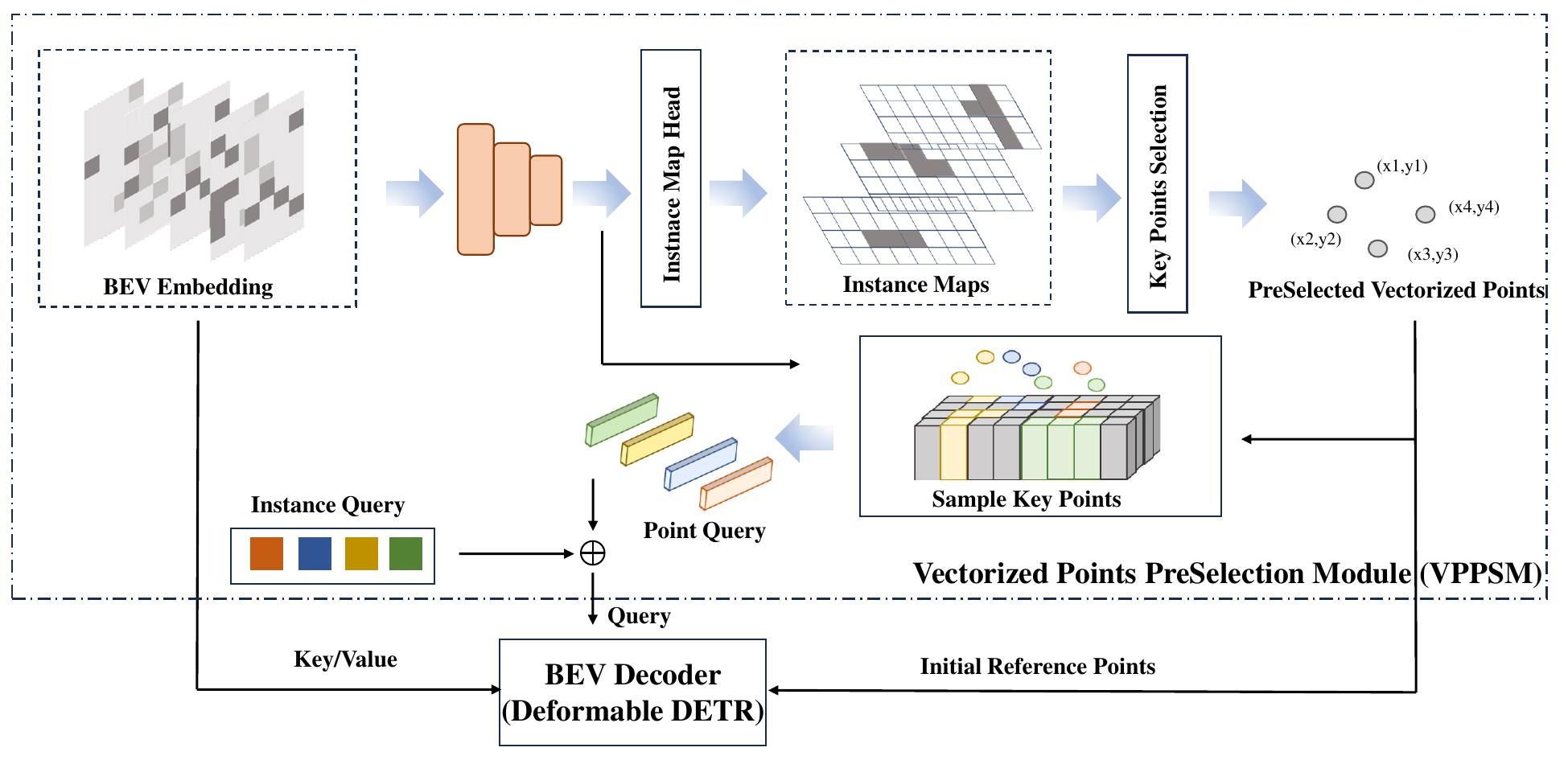}
      \caption{Diagram of the Vectorized Points PreSelection Module (VPPSM). In simple terms, an instance map is learned from BEV features. The coarse vector points are selected in each instance map according to the key point sampling principle. The instance features and the geometric positions of these selected vector points are encoded as a point query.}
      \label{Fig.3}
      \vspace{-1.7em}
\end{figure*}

\subsection{Framework}
\label{sec:framwork}
As shown in Fig.~\ref{Fig.2}, DTCLMapper follows the basic framework of constructing a vectorized map, comprising an image backbone, view transformer, BEV decoder, and multi-task head.
Multi-view images $I_j$ $(j \in [0,n])$ are encoded by a pre-trained backbone, such as ResNet (Res)~\cite{restnet}.
Then the deep feature of each image is aggregated into a BEV embedding 
$F_{bev}=Prob(I_j,VT)$ $(F_{bev}\in\mathbb{R}^{H{\times}W{\times}C})$ ($H$ and $W$ are the height and width of a BEV embedding, respectively, and $C$ denotes the learning dimension) through View Transformer (VT). 
As this paper focuses on dual temporal learning rather than proposing a view transformer architecture, 
we choose the classic method as the view transformer module, 
BEVFormer ($Prob_B$)~\cite{BEVFormer} and LSS ($Prob_L$)~\cite{LSS}.

\begin{equation}
    Z_c * \left[\begin{array}{l} u \\ v \\ 1\\ \end{array} \right] = T_{in} \cdot \left[\begin{array}{l} X_c \\ Y_c \\ Z_c\\ \end{array} \right]
    \left[\begin{array}{l} X_c \\ Y_c \\ Z_c\\ 1\\ \end{array} \right] = T_{ex} \cdot \left[\begin{array}{l} X_w \\ Y_w \\ Z_w\\ 1\\ \end{array} \right],
    \vspace{-0.15cm}
    \label{eq0}
\end{equation}

\begin{equation}
    Prob_{B} = \sum_{i=0}^{8} F_A(Res(I_j), P3(T_{in}, T_{ex},[X_w, Y_w, Z_i]^{\rm T})),
\end{equation}
\begin{equation}
    Prob_{L} = F_B(Res(I_j),P2(T_{in}, T_{ex},[u, v]^{\rm T},{Z_{c}^{p}})),
\end{equation}
where $T_{in}$ and $T_{ex}$ are intrinsic and extrinsic matrixes of each camera. $X_w$, $Y_w$ and $Z_i$ are in the ego coordinate system. $X_c$, $Y_c$, and $Z_c$ are in the camera coordinate system. $u$ and $v$ are the pixel coordinate system. ${Z_{c}^{p}}$ is the value of predicted depth. $P3$ is VT from 3D points to 2D images. $F_A$ is deformable cross-attention learning at 2D features to obtain corresponding 3D features. $P2$ is VT from 2D pixels to 3D points. $F_B$  is the projection of 2D features to corresponding 3D points based on ${Z_{c}^{p}}$.  
And the Deformable DETR~\cite{ddetr} is employed for the BEV decoder. The input of decoder contains a BEV embedding, an instance query $Q\in \mathbb{R}^{N{\times}C}$($N$ is the number of instances), the initial reference points $P\in \mathbb{R}^{N{\times}N_p{\times}2}$ ($N_p$ is the fixed number of vector points in each instance) and the point query $Q_{P} \in \mathbb{R}^{N{\times}N_p{\times}C}$. 
Finally, linear layers combined with LayerNorm and ReLU are used as the task head for predicting the label and box of the instance $Ins_i$.

Within this basic framework, two layers of consistency learning, namely Instance Consistency Learning (ICL) and Map Consistency Learning (MCL), are seamlessly distributed at the instance and map layers, respectively. 
At the ICL layer, to improve the quality of vector points, we move away from the random initialization style previously used in MapTR and first design a VPPSM. 
The initial vector points are selected from coarse instance maps. By mixing the semantic and positional features of each vector point, we obtain a point query, which is then optimized through multiple layers in the decoder. 
To enhance the expressiveness of the instance feature aggregated from VPPSM, we start from the geometric alignment and semantic categorization between temporal instances, designing a consistency learning approach for the instance features based on the concept of contrastive learning. 
Moreover, at the MCL layer, the approach involves comparing the geometry consistency of global instances in grid maps translated from vectorized HD maps, achieved through a newly introduced Map Occupancy Loss (MO Loss).

\subsection{Temporal Instance Consistent Learning}
\label{sec:il}
As mentioned above, most previous works~\cite{BEVFormer,BEVFormerv2,beverse}, use temporal information to improve the richness of visual features in BEV space. 
In their works, global BEV features, as the basic unit, are analyzed through convolution or attention mechanism to get the global weight matrix in temporal frames. Then, this matrix is leveraged to fuse the missing features of this frame from the historical BEV features. 
During the BEV feature self-learning process from scratch, minimal differences in static objects across consecutive frames within short temporal sequences (as analyzed at the top of Fig.~\ref{Fig.intro}) hinder the extraction of reliable deep features from perspective images via existing methods. Specifically, subtle differences can guide modules to repetitively learn overlapping features from similar characteristics, potentially causing feature redundancy phenomena.
If maps are decoded on this redundant feature, truly reliable instance features will be lost in each frame. 
Therefore, we consider that temporal fusion information is beneficial for dynamic target detection, but it lacks the ability to merge static map elements. 

To address this issue, we propose an Instance Consistent Learning (ICL) strategy in this section. 
ICL analyzes the differences between map instances themselves from temporal consistency and compensates for the differences by the unique space of different semantics. 
Instead of merging BEV features globally, this focuses on the feature of each instance itself.
Precisely, as more discriminative instance features are helpful in improving the quality of temporal instance consistency learning, VPPSM (Sec.~\ref{sec:pre}) is firstly leveraged to improve the instance feature from the vector points. Then, a novel contrastive model, AIFCL (Sec.~\ref{sec:fea}), is used to make the associated instance feature between temporal more closely with the help of negative samples. 

\subsubsection{Vector Point PreSelection Module}
\label{sec:pre}
At present, map element instances in a vectorized map are decoded from the BEV feature through queries of the DETR architecture. Each initial query is not only randomly generated, but also shares a learning dimension for all sampling points. 
After large-scale data training, an initialization point query with the largest common factor as shadow will be generated. 
However, in terms of instance queries whose vector points are far from the common points as a shadow, more iterations are needed for regression learning. 
Thus, this is an inefficient initialization for an HD map with various shape instances.

Therefore, intending to efficiently aggregate instance features, we design a Vector Point PreSelection Module (VPPSM) to achieve differentiated initialization. 
As shown in Fig.~\ref{Fig.3}, given the BEV feature map $F_{bev}\in \mathbb{R}^{H{\times}W{\times}C}$, our method generates the coarse instance feature $F_{pos} \in R^{H{\times}W{\times}C}$ about vector points. 
A simple head is used to obtain instance maps $Map_{inst} \in R^{H{\times}W{\times}N}$, where $N$ is the set number of preselected instances. 
Then, the instance maps, represented by a form of the grid map, can be constructed:
\begin{equation}
    F_{pos} = \theta_p(F_{bev}), 
\end{equation}
\begin{equation}
    Map_{inst} = \theta_i(F_{pos}),
\end{equation}
where $\theta_p$ is well-established ResNet structure with three layers and $\theta_i$ is a head structure. 
Note that the instance map prediction follows a supervised learning fashion. 
The details of the learning supervision are presented in Sec.~\ref{sec:loss}. 
 
The vector points are used to describe the geometric structure of an instance, meaning that their positions are within the range occupied by the instance. Then we propose a weight module for learning the importance of key points. 
The key points with high weights are the initial vector points of which the geometric positions are output as the preselected vector points $coord_{xy}\in N{\times}P{\times}2$:

\begin{equation}
\begin{aligned}
    W_{map} &= H_w(Map_{inst}),\\
    coord_{xy} &= Norm(TopK(Flatten(W_{map})),\\
\end{aligned}
\end{equation}
where $H_w$ is the weight module and is implemented through a 1D convolution. 
$TopK$ follows a pattern to select the position with the highest score and outputs the position coordinates. 
$Norm$ is used to achieve coordinate normalization and avoid inconsistent coordinate systems. 
The deep instance features of these points $pos_{insfea}$ are obtained by back-projecting $S_a$ them onto the initial position feature map:

\begin{equation}
\begin{aligned}
    coord_{xy}^{'} &= coord_{xy} \cdot [H,W]^{\rm T},\\
    coord_{xy}^{''} &= coord_{xy}^{'}[0]+coord_{xy}^{'}[1]*W,\\
    pos_{insfea} &= S_a(Flatten(F_{pos}),coord_{xy}^{''}).
\end{aligned}
\end{equation}

Though $pos_{fea}$ includes deep instance features, it lacks relative geometric position features, which is an indispensable knowledge of vector points. 
Thus, we leverage two MLP layers $M_g$ to encode the normalized pixel coordinates of the vector points in the instance map $coord_{xy}$. Then the geometric position features of these can be calculated:
\begin{equation}
    pos_{geofea} = M_g(coord_{xy}).
\end{equation}

Finally, the preselected vector points $coord_{xy}$ and the feature $pos_{fea}{=}pos_{insfea}{+}pos_{geofea}$ are used as the initial reference point $P_m$ and the point query $Q_{P_m} \in \mathbb{R}^{N{\times}N_p{\times}C} $ to decode each map instance. 
Note that the query $Q{=}Q_m{+}Q_{P_m}$ is sent into the deformable DETR where $Q_m \in \mathbb{R}^{N{\times}C}$ is initialed as a random variable and extended to be consistent with $Q_{P_m}$ by repeat operation.   

\begin{algorithm}[tb]  
  \caption{Selection and Learning of Positive and Negative Samples:} 
  \label{code}  
   \textbf{Input:} \\
      $HS_0,HS_1,...,HS_n$: History instance feature;\\
      $CS_0,CS_1,...,CS_n$: The score of current instances; \\
      $CF_i$: The contrastive feature of now instances; \\
      $B_i$: The box value of now instances; \\
    \textbf{Output:} 
       $Loss_{cst}$ 
    \begin{algorithmic}[1]
    \STATE def FindBox (singlebox, boxset, label):\\ 
    \COMMENT{Find the nearest box number from this boxset to this singlebox in the same label.}
    \STATE def FindNeg (set, label):\\ 
    \COMMENT{Find the number without this label in this set.}
    \FOR{$i = 0,\dots, n$}
        \STATE $N_0 = HigherCmp(CS_i,CS_{N_0})$
        \STATE $N_1 = HigherCmp(CS_i,CS_{N_1})$
        \STATE $N_2 = HigherCmp(CS_i,CS_{N_2})$
    \ENDFOR
    \FOR {$i = 0,\dots,2$}
        \STATE $pos_i= FindBox(B_{N_i},(HS_0,HS_1,...,HS_n),i)$
        \STATE $neg_i= FindNeg((HS_0,HS_1,...,HS_n), i)$
        \STATE $v= CF_{N_i}$
        \STATE $l_i=\sum_{neg}exp⁡(v*neg_i-v*pos_i)$
    \ENDFOR
    \STATE $Loss_{cst} = log[1+\sum_{i=0}^{2}l_i]$
   \end{algorithmic}
\end{algorithm}

\subsubsection{Contrastive Learning of Aggregated Instance Feature}
\label{sec:fea}
The instance feature aggregated from VPPSM is the output of single-frame perception information. 
As the instance features extracted from occluded scenes in a single frame are extremely sparse, these weak features can be easily overlooked during learning.
Thus, the weak instance features need more attention to improve the quality of an HD map.
At the same time, we observe a phenomenon that the temporal context priors of instances differ due to viewing angles. 
Further consideration, is whether this temporal difference can enhance the richness of weak instance features and how to link this difference. 

Based on these considerations, we design a contrastive learning model of aggregated instance features, starting from the consistency of these features in their respective temporal spaces to link instance features with different differences. 
As represented in \textbf{Algorithm~\ref{code}}, the contrastive learning model is based on the history instance feature and the current instance feature, developing the strategy of consistent learning with the help of the instance box, the label and the instance score that is under the supervision. 
Considering the unique clustering space between different labels, the selection of positive and negative samples will be carried out independently under different labels, ensuring a broad space for contrastive learning.

More specifically, for the current frame, the aggregated instance feature with a higher instance score in each semantic label is sent to the contrastive head. This score is generated on the comprehensive evaluation of instance and location accuracy under the supervision of ground truth. 
And then `$v$' embedding can be obtained, as the value in contrastive learning. 
Note that similar to the work of~\cite{idol}, an extra light-weight FFN is used as a contrastive head to generate the contrastive embedding.
Since the number of queries is larger than the number of instances existing in each frame, the number of `$v$' is designed as a dynamic norm. 
For each label, a fixed maximum amount of `$v$' is set. 
The setting of this maximum value has been discussed in experiments to avoid polluting the space with low-quality repeated instances. 
The final number is determined by the joint constraint of the set maximum number and the actual number of detected instances in this frame.

Subsequently, an elaborate selection mechanism is served for positive and negative samples. 
In adjacent frames of temporal series, changes in the dynamic environment may cause differences in features between instances with a similar position. 
In particular, parts of instance features are sparse due to the occlusion. However, through the learning process of positive samples, sparse feature expression can be enhanced from relatively rich features.
Thus, we choose historical instances $Ins_{his}$ within a certain range around the `$v$' instance as positive samples $Sam_{pos}$.
The negative samples $Sam_{neg}$ are selected from the negative label, allowing the features of instances with the same labels to be pulled closer together. The negative label refers to a category other than the label of `$v$' itself.
The number of negative samples with each negative label is fixed. 
Several of the highest instance scores in the current frame with negative labels are selected as negative samples to improve learning efficiency.

\begin{equation}
    \left\{
        \begin{array}{ll}
            L_{p},& label(Ins_{his}) = label(v)\\
            L_{n},& label(Ins_{his}) != label(v),
        \end{array}
    \right.
\vspace{-0.1em}
\end{equation}
\begin{equation}
    Sam_{pos} = min_{i}(Sqrt(X_{L_{p_i}} - X_{v})^2+(Y_{L_{p_i}}-Y_{v})^2)),
\vspace{-0.5em}
\end{equation}

\begin{equation}
    Sam_{neg} = \mathop{max}\limits_{k}(L_0, \ldots L_{n}),
\vspace{-0.3em}
\end{equation}
where $L_{p}$ and $L_{n}$ are chosen by the label. $k$ is a fixed setting. Finally, the contrastive loss of each class is defined as follows:
\begin{equation}
\begin{aligned}
\label{eq1}
    Loss_{cst}^{'}= -log \frac{exp⁡(v*k^+)}{exp⁡(v*k^+)+\sum_{k^-}exp⁡(v*k^-)}\\
           =log[1+\sum_{k^-}exp⁡(v*k^--v*k^+)],
\end{aligned}
\end{equation}
where $k^+$ and $k^-$ are positive and negative samples, respectively. Then we extent Eq.~\ref{eq1} into Eq.~\ref{eq2} based on all classes:
\begin{equation}
\label{eq2}
    Loss_{cst}=log[1+\sum_{c}\sum_{k^+}\sum_{k^-}exp⁡(v_c*k^--v_c*k^+)],
\end{equation}
where $v_c$ denotes the  `$v$' embedding in each class.

\subsection{Temporal Map Consistent Learning}
\label{sec:ml}
The ICL, presented in Sec.~\ref{sec:il}, reinforces the expression of instances from a semantic layer. 
To further enhance HD map learning, we consider that there is also a wealth of information about the geometric position relationships between instances. 
More importantly, since instances are static in a map, the geometric positions of instances between temporal frames should be consistent, providing a natural condition for the global position constraints of instances. 
Thus, we further propose consistency learning based on the geometric position of global instances to enhance the generalization performance from a global view. 
A grid map represents the geometric position of instances. The reason lies in that the grid map is expressed consistently for all instances, that is, the occupancy state of location, providing an effective basis to implement global consistency of temporal instances.

The remaining question is how to translate a vectorized map into a grid map. An instance composed of vector points $P_m$ is the basic unit of a vectorized map. Generally, points with directional information, which are supervised in this work, can be connected to a vector line. It means each instance can be converted into a polyline which can be smoothly projected onto the grid plane:  
\begin{equation}
    V2GMap = Raster([Poly(P_1),...,Poly(P_{N^{'}})]),
\end{equation}
where $Poly$ means a polyline and $N^{'}$ ($N^{'}$ is less than $N$) is the number of dependable instance. 
It should be noted that only an instance with a normal confidence level will be projected into the grid map. 
This confidence level is dependent on the predicted probability of the instance class, rather than the supervision of ground truth. 
The main drivers fuelling this setting are the expected benefits of the HD map with greater generalization ability.

Conventionally, random grids in temporal grid maps aligned by the ego motion have a similar occupancy probability. Under the guidance of this idea, an MO Loss is designed to control the consistency between temporal grid maps. Given the coordinates $W_{xy}{=}{[X_0,Y_0,0],\ldots,[X_w,Y_w,0]}$ in the world coordinate system, the process of aligning is achieved through the motion transformation, as explained in Eq.~\ref{tr} and Eq.~\ref{tr1}. 

\begin{equation}
\label{tr}
   XY^{'} = [R|T]_{3\times3} \cdot W_{xy},
\end{equation}
\vspace{-1.0em}
\begin{equation}
\label{tr1}
\begin{aligned}
    V2GMap_{t-i}^{'} = S_a(Flatten(V2GMap_{t-i}),XY^{'}),
\end{aligned}
\end{equation}
\noindent where $R$ and $T$ are rotation and translation for the ego, respectively. $S_a$ is the sampling function, according to $XY^{'}$ sampling points in the $V2GMap_{t-i}$ to obtain the occupied state, which is consistent with the previous definition. Finally, the occupancy loss is defined as Eq.~\ref{oc}:
\begin{equation}
\label{oc}
    Loss_{ol}=\sum_{i=1}^{m}L(V2GMap_{t-i}^{'},V2GMap_t),
\end{equation}
where $L$ is L1 loss function and $m$ is the number of sequence temporal frames.

\begin{table*}[t]
\fontsize{9}{13.8}\selectfont
\renewcommand{\arraystretch}{1.1}
\setlength\tabcolsep{9pt}
\caption{Results on the nuScenes dataset. $*$ represents that the data is realized at the same implementation by open source.}
\vspace{-2.0em}
\label{tab:nus_sota}
\begin{center}
\resizebox{1.0\linewidth}{!}{\begin{tabular}{lcccccccc}
\toprule [2pt]
Method          & Backbone & View-Trans & Layer nums & Epoch & $\rm{AP_{divider}}$         & $\rm{AP_{ped}}$         & $\rm{AP_{road}}$         & mAP           \\ \midrule [1pt] 
HDMapNet~\cite{hdmapnet}        & EB0 & MLP        & -          & 30    & 28.3          & 7.1           & 32.6          & 22.7          \\
VectorMapNet~\cite{vecmapnet} & R50 & IPM        & -          & 110   & 47.3          & 36.1          & 39.3          & 40.9          \\
InstaGram~\cite{instmap}      & EffiNet-B0 & MLP   & -             & 30  &40.8	&30.0	&39.2	&36.7                              \\
MapTR~\cite{maptr}           & R50 & GKT  & 1          & 24    & 51.5          & 46.3          & 53.1          & 50.3          \\
MapVR~\cite{vect2raster}    & R50 & GKT  & 1          & 24    & 54.4          & 47.7          & 51.4          & 51.2         \\
PivotNet~\cite{pivotnet}        & R50 &BEVFormer  & 4          & 30    & 56.5          & 56.2          & 60.1          & 57.6          \\
MapTRv2*~\cite{maptrv2}        & R50 & LSS        & -          & 24    & 61.4          & 57.8          & 60.4          & 59.9          \\
\rowcolor[gray]{.9}
DTCLMapper-b    & R50 & GKT & 1          & 24    & 53.9  & 51.2  & 55.8  & 53.6         \\

\rowcolor[gray]{.9}
DTCLMapper-l    & R50 & LSS        & -          & 24    & \textbf{62.6} & \textbf{59.7} & \textbf{63.4} & \textbf{61.9} \\ \bottomrule [2pt]
\end{tabular}}
\end{center}
\vspace {-1.0em}
\end{table*}
\begin{figure*}[htb]
      \centering
      \includegraphics[scale=0.4]{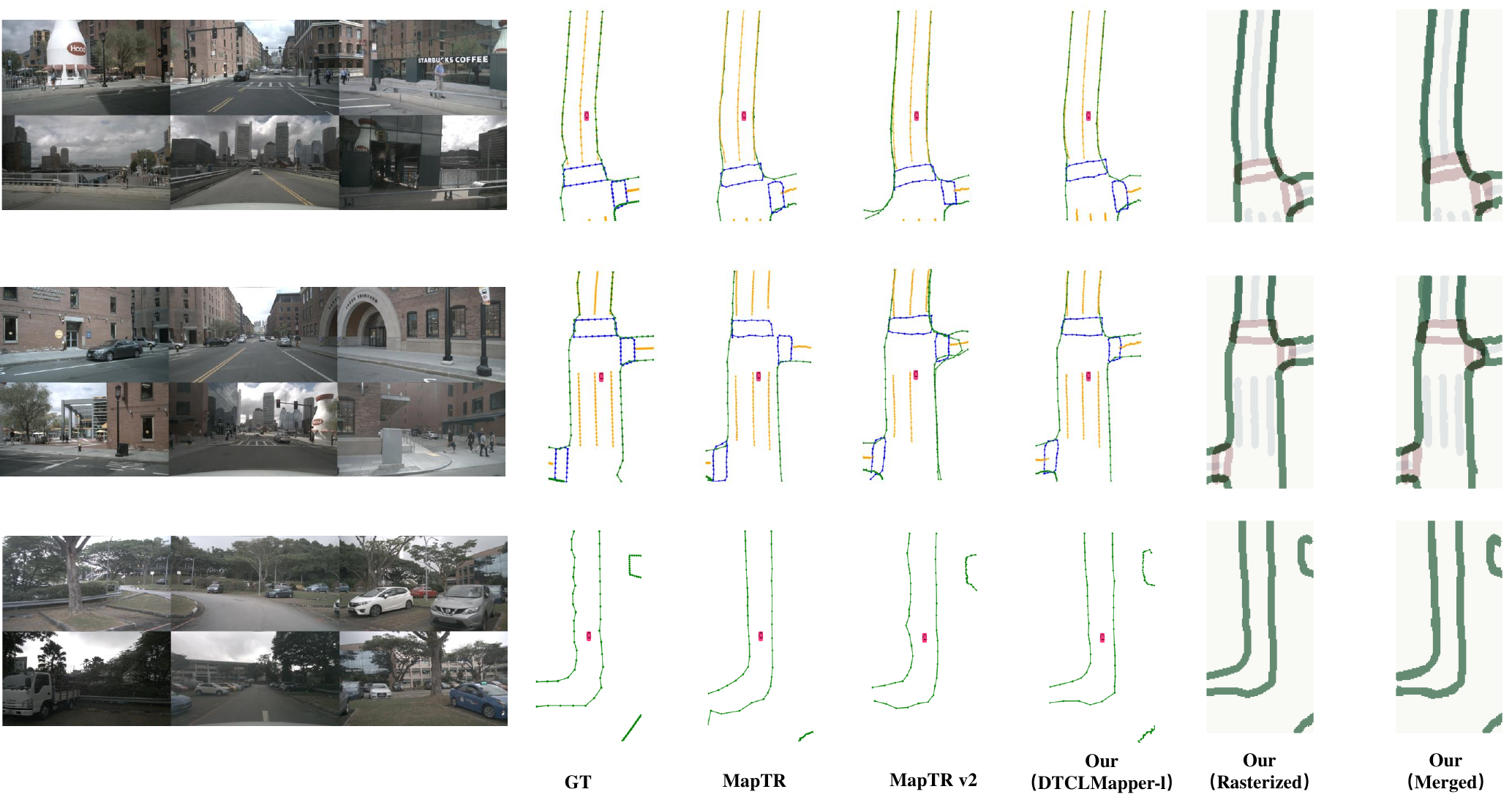}
      \caption{Visualization on the nuScenes validation dataset. From left to right are multi-view perspective images, GT, MapTR~\cite{maptr}, MapTRv2~\cite{maptrv2}, and our work. The last two columns are grid maps rasterized from the vectorized map and grid maps merged from temporal grid maps.}
      \label{fig.nus}
      \vspace {-1.0em}
\end{figure*}

\subsection{Training Loss}
\label{sec:loss}
The training loss includes several basic losses, classification loss, point loss, and direction Loss, which are based on the two matchings. In addition, instance map loss works in VPPSM. $Loss_{cst}$ and $Loss_{ol}$ are described above.

\textbf{Matching:} It is divided into two layers, the instance and the point. 
Following DETR, the Hungarian algorithm is utilized for the instance. For a pair of the matched instances, vector points are matched based on the Manhattan distance.   

\textbf{Classification Loss:} Each predicted instance $p_i$ is assigned a class label $g_i$ through the instance matching. Focal loss is used to measure this:
\begin{equation}
    Loss_{cls} = \sum_{i=0}^{N-1}L_{focal}(p_i,g_i).
\end{equation}

\textbf{Point Loss:} A pair of the matched instances, the predicted point set $v_{ij}$ is assigned with a GT point set $vg_{ij}$ after the point matching. This loss realized by L1 loss:
\begin{equation}
    Loss_{pts} = \sum_{i=0}^{N-1}\sum_{j=0}^{M-1}L_{1}(v_{ij},vg_{ij}).
\end{equation}

\textbf{Direction Loss:} Based on the point matching, this considers connecting lines between adjacent points. Supervision is done by the cosine similarity of the connecting lines $e_{ij}$ and $eg_{ij}$: 
\begin{equation}
    Loss_{dirs} = \sum_{i=0}^{N-1}\sum_{j=0}^{M-2}L_{cosine\_similarity}(e_{ij},eg_{ij}).
\end{equation}

\textbf{Instance Map Loss:} We follow the work of~\cite{hdmapnet}, the loss is computed by:
\begin{equation}
    Loss_{var} = \frac{1}{C}\sum_{c=1}^{C}\frac{1}{N_c}\sum_{j=1}^{N_c}[||\mu_c - f^{instance}_j||-\delta_v]^2_+,
\end{equation}
\begin{equation}
    Loss_{dist} = \frac{1}{C(C-1)}\sum[2\delta_d-||\mu_{cA}-\mu_{cB}||]^2_+,
\end{equation}
where $C$ is the number of instances in GT, $N_c$ is the number of the predicted instances, $\mu_c$ is the mean of a predicted instance, $||.||$ is L2 norm, and $[x]_+$ denotes the element maximum compared with zero.

\textbf{Overall Loss:} The overall loss is computed by:
\begin{equation}
\begin{aligned}
    Loss = \lambda_1 Loss_{cls} + \lambda_2 Loss_{pts} + \lambda_3 Loss_{dirs} + \\\lambda_a Loss_{cst} + \lambda_b Loss_{ol} + \lambda_c Loss_{var} + \lambda_d Loss_{dist},
\end{aligned}   
\end{equation}
where $\lambda_1$ to $\lambda_3$ are the loss weights of detecting instances. $\lambda_a$ to $\lambda_d$ are the corresponding loss weights in ICL and MCL.

\section{Experiment}
\begin{table}[tb]
\fontsize{9}{11.8}\selectfont
\renewcommand{\arraystretch}{1.1}
\caption{Results on the Argoverse dataset. $*$ represents that the data is realized at the same implementation by open source.}
\vspace {-1.0em}
\label{tab:arg_sota}
\begin{center}
\resizebox{0.95\linewidth}{!}{\begin{tabular}{lccccc}
\toprule [2pt] 
Method                               & Epo                    & $\rm{AP_{d}}$         & $\rm{AP_{p}}$         & $\rm{AP_{r}}$                                 & mAP                                    \\ \midrule [1pt] 
HDMapNet~\cite{hdmapnet}                             & 6   &5.7	&13.1	&37.6	&18.8 \\
VectorMapNet~\cite{vecmapnet}                      & 24  &6.1	&38.3	&39.2	&37.9 \\
MapVR~\cite{vect2raster}                                & 6   &60.0	&54.6	&58.0	&57.5     \\
MapTR*~\cite{maptr}                                & 6   &62.7	&55.0	&58.5	&58.8     \\
MapTRv2*~\cite{maptrv2}                             & 6     & 68.8  & 61.3   & 63.4      & 64.5                                  \\
\rowcolor[gray]{.9}
DTCLMapper-b & 6 & \textbf{64.7} & \textbf{56.3} & \textbf{60.1} & \textbf{60.4}  \\ 
\rowcolor[gray]{.9}
DTCLMapper-l & 6 & \textbf{69.4} & \textbf{61.9} & \textbf{64.1} & \textbf{65.1}  \\ \bottomrule [2pt]
 
\end{tabular}}
\end{center}
\vspace {-1.0em}
\end{table}

\subsection{Datasets}
\textbf{nuScenes}~\cite{nus} is a dataset for autonomous driving. In the nuScenes dataset, the surround-view six cameras can provide panoramic multi-view information. It has global BEV HD maps which can supervise the framework effectively. Moreover, this continuous dataset has complete temporal information, which is the core of the task.

\textbf{Argoverse 2}~\cite{argoverse} is usually used to study for perception and forecasting. 
It has seven ring cameras and two stereo cameras. We only use the seven ring cameras to construct maps. 
This dataset has comprehensive BEV HD maps of two cities, ensuring the diversity and capacity to supervise learning for semantic mapping. 

\subsection{Evaluation Metrics}
Similar to previous works on vectorized map construction~\cite{vecmapnet,maptr}, our work chooses three map elements to evaluate, covering lane divider, pedestrian crossing, and road boundary. Based on Chamfer Distance (CD), the Average Precision (AP) is used as the evaluation metric. AP is calculated under three CD thresholds $\{0.5m,1.0m,1.5m\}$ and then is averaged as the final evaluation metric (mAP). 
In a local map, the detection range of map elements is $60m$ in the $x$ direction and $30m$ in the $y$ direction.

\subsection{Implementation Details}
All experiments are conducted with NVIDIA RTX A6000 GPUs. 
The training batch is $16$ and the initial learning rate is $1.5e^{-3}$. 
The model is trained $24$ and $6$ epochs on the nuScenes and Argoverse datasets respectively.
The resolution of grid maps rasterized from vectorized maps is $0.15m$. The size of maps is $(400,200)$ on the nuScenes dataset and $(200,400)$ on the Argoverse dataset. 
The rasterized threshold of an instance is $0.4$.
The weight of loss is set as, $\lambda_1{=}2$, $\lambda_2{=}5$, $\lambda_3{=}0.005$, $\lambda_a{=}0.1$, $\lambda_b{=}1$, $\lambda_c{=}1$, $\lambda_d{=}0.1$. 
Other training settings are consistent with MapTR~\cite{maptr}.

\subsection{Comparison with State-of-the-Arts}
We compare our works with other previous HD map construction methods. \textbf{HDMapNet}~\cite{hdmapnet} is the first work to detect instances in an HD map. \textbf{VectorMapNet}~\cite{vecmapnet} introduces an end-to-end  network to build vectorized maps. \textbf{MapTR}~\cite{maptr} brings map construction to a real-time level. 
\textbf{MapVR}~\cite{vect2raster} has also investigated constructing rasterized maps from vectorized maps. 
Recently, \textbf{PivotNet}~\cite{pivotnet} and \textbf{MapTRv2}~\cite{maptrv2} appear in the map generation competition one after another. 
As the view transformation methods of these methods are inconsistent, \textit{e.g.}, MLP~\cite{mlp}, IPM~\cite{ipm}, transformers, and LSS~\cite{LSS}, the subsequent comparison are completed based on the same transformation methods.

\begin{figure}[tb]
      \centering
      \includegraphics[scale=0.38]{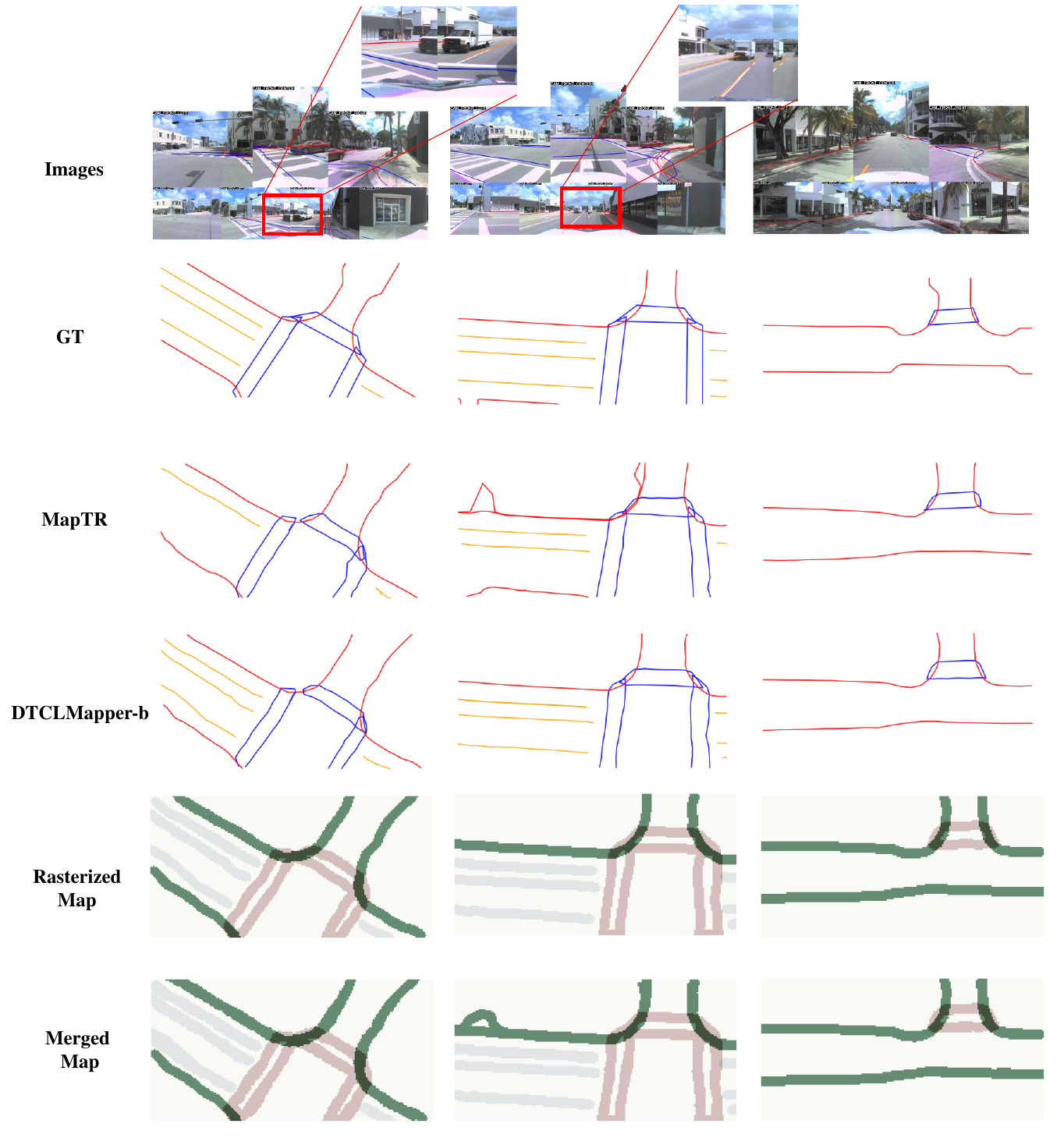}
      \caption{Visualization on the Argoverse validation dataset. From top to down are multi-view perspective images, GT, MapTR~\cite{maptr}, and the proposed DTCLMapper-b. The last two lines are also grid maps rasterized from the vectorized map and grid maps merged from temporal grid maps.}
      \label{fig.arg}
      \vspace{-1.5em}
\end{figure}

\textbf{Results on nuScenes:}
Table~\ref{tab:nus_sota} shows the comparison results on the nuScenes dataset. 
To conduct a comprehensive comparison, we verify our methods in two different transformation manners, \textit{i.e.}, DTCLMapper-b using a single layer of transformer, and DTCLMapper-l using LSS. 
Compared to the transformer-based MapTR and MapVR, the mAP ($53.6\%$) of DTCLMapper-b is higher, 
yielding respective ${+}3.3\%$ and ${+}2.4\%$ gains. While using LSS, our DTCLMapper-l model can outperform LSS-based MapTRv2 and obtain the highest mAP score of $61.9\%$.  
Compared to other models~\cite{hdmapnet,vecmapnet,instmap} learning instance features from a single frame, our DTCLMapper obtains significant gains 
thanks to the proposed methods enhancing the representation of weak instance features through temporal consistency. 
Apart from quantitative results, we further conduct qualitative analysis.
The visualization results are shown in Fig.~\ref{fig.nus}. 
Compared to MapTR and MapTRv2 methods, our method performs more consistently in different driving scenarios and yields accurate HD maps. 
Additionally, the grid maps rasterized from the vectorized maps (resolution of $0.15m$) and a merged map fused from multi-frames are also presented to prove the validity of the rasterization method.

\begin{figure*}[htb]
      \centering
      \includegraphics[scale=0.3]{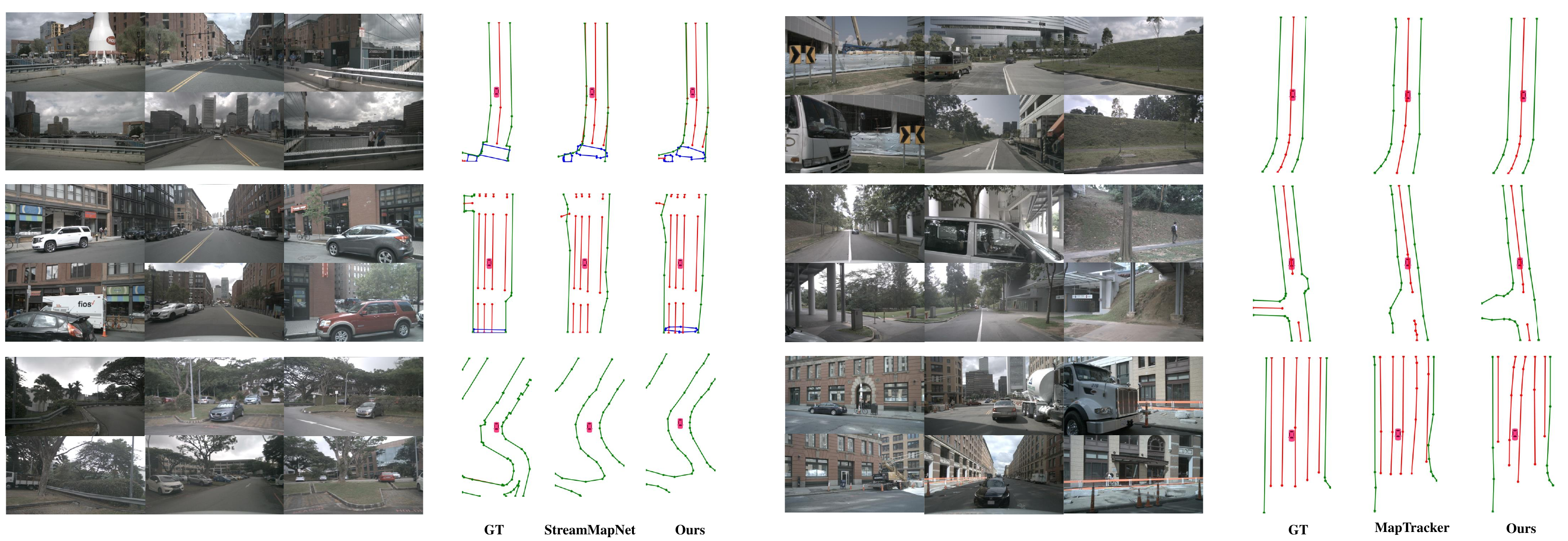}
      \vspace{-0.25cm}
      \caption{Visualization of long temporal fusion on the nuScenes validation dataset. On the left is the comparison against StreamMapNet~\cite{streammapnet}. On the right is the comparison against MapTracker~\cite{maptracker}.}
      \label{fig.nus_st}
      \vspace {-1.0em}
\end{figure*}
\begin{table}[tb]
\fontsize{9}{11.8}\selectfont
\renewcommand{\arraystretch}{1.1}
\caption{Results on the different memory buffer strategies. $*$ represents that the data is realized at the same implementation by open source. }
\vspace {-1.0em}
\label{tab:new_data}
\begin{center}
\resizebox{0.95\linewidth}{!}{\begin{tabular}{lcccc}
\toprule [2pt] 
Method    & Epo        & training bsz            & mAP  &FPS \\ \midrule [1pt] 
StreamMapNet*~\cite{streammapnet}         & 30   	&8		&60.3 &16.3 \\
\rowcolor[gray]{.9}
DTCLMapper-Stream                     & 30   	&8		&61.4 &15.9 \\
MapTracker*~\cite{maptracker}        & 30   	&(4,16,8)		&65.4  &12.8    \\
\rowcolor[gray]{.9}
DTCLMapper-Tracker              & 30  	&(4,16,8)		&66.9  &12.3 \\   
\bottomrule [2pt]
 
\end{tabular}}
\end{center}
\vspace {-2.0em}
\end{table}
\begin{table}[tb]
\large
\fontsize{9}{12.8}\selectfont
\begin{center}
\caption{Ablation results on proposed modules.}
\vskip -1ex
\label{ablation}
\begin{tabular*}{0.45\textwidth}{@{\extracolsep{\fill}}ccccccc}
\toprule [2pt]  
VPPSM & AIFCL & MCL  & PE  & mAP  &FPS \\ \midrule [1pt] 
    &   &   &  & 50.3 &16.5 \\
 \checkmark   &    &   &   & 51.6  &16.1 \\
 \checkmark   & \checkmark  &  &   &52.5 &16.1  \\
 \checkmark   & \checkmark  & \checkmark &  & 52.9 &16.1 \\ 
 \checkmark   & \checkmark  & \checkmark & \checkmark & 53.6 &16.0 \\ \bottomrule [2pt]
\end{tabular*}
\end{center}
\vspace {-2.0em}
\end{table}

\textbf{Results on Argoverse:}
Table~\ref{tab:arg_sota} shows the comparison results on the Argoverse dataset. In line with previous experiments, we test our models with transformer-based (DTCLMapper-b) and LSS-based (DTCLMapper-l) view transformations, respectively. Consistently, our work outperforms state-of-the-art approaches in all classes, obtaining the best mAP score of $65.1\%$. It demonstrates that the proposed dual consistent learning seamlessly adapts to various data.
Fig.~\ref{fig.arg} shows the visualization of MapTR and DTCLMapper from the validation set. The figure also represents the comparison results in the case of partial occlusion, where our method can accurately detect the instance location, proving a higher sensitivity for instances in DTCLMapper. The consistent improvements in quantitative and qualitative comparisons showcase the effectiveness of our proposed temporal consistency methods.

\begin{figure}[t]
      \centering
      \includegraphics[scale=0.33]{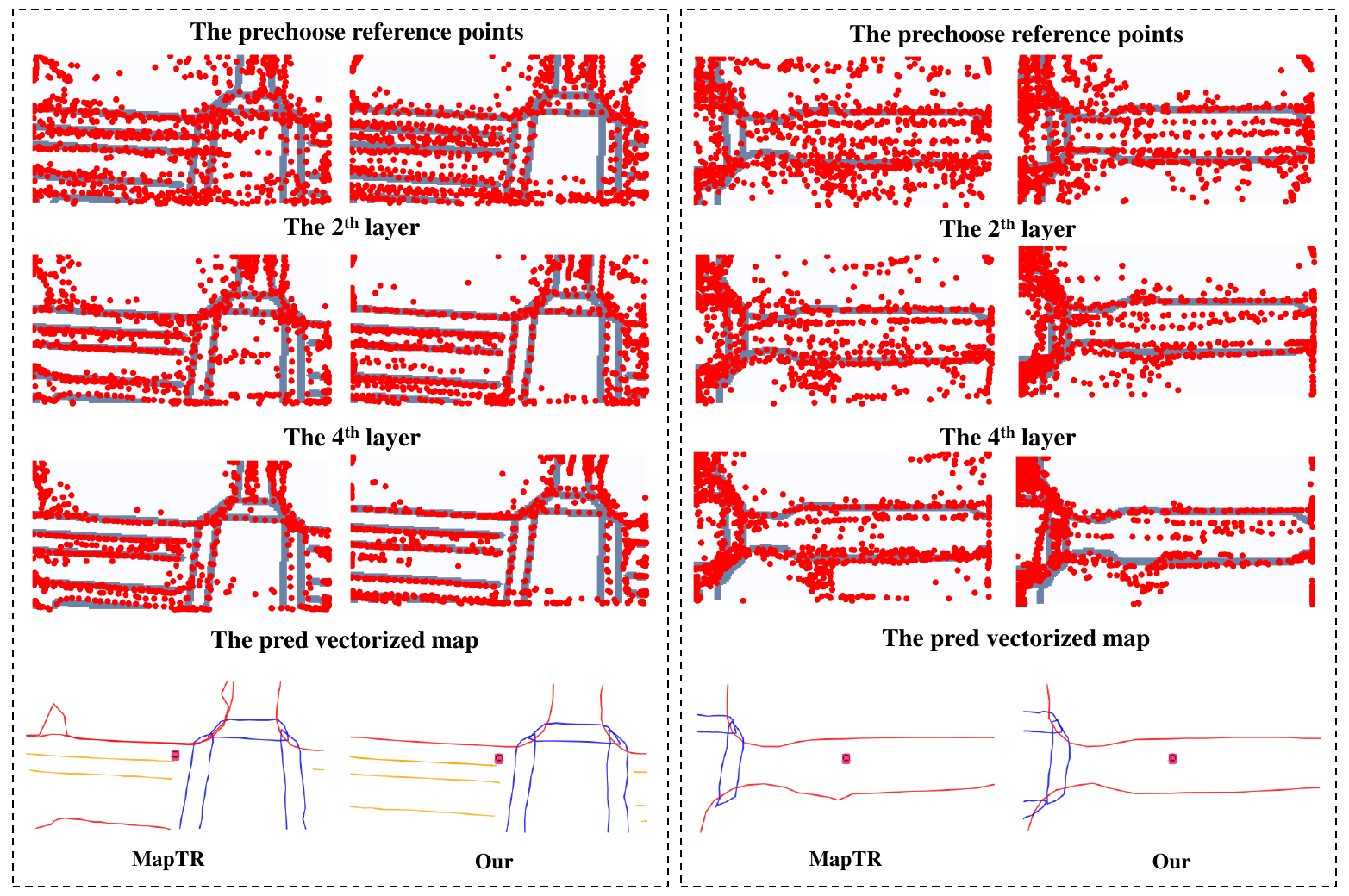}
      \vspace {-1.0em}
      \caption{Visualization of vector points regression. Two methods are compared, random initialization in MapTR~\cite{maptr} and the proposed VPPSM.}
      \label{fig.prechoose}
      \vspace {-1.5em}
\end{figure}

\textbf{Results of Different Memory Buffer Strategies: }
Table~\ref{tab:new_data} shows the results of different memory buffer strategies on the nuScenes dataset. 
Two methods are chosen, including StreamMapNet~\cite{streammapnet} and MapTracker\cite{maptracker}. The proposed temporal consistency learning contributes to improving the mapping quality of these two existing temporal memory buffer strategies, achieving scores of $61.0\%$ and $66.9\%$ in mAP, respectively. Moreover, in terms of efficiency, we remain at a comparable level with these methods.
Fig.~\ref{fig.nus_st} further demonstrates the beneficial effects of our proposed method on map generation, where our approach accurately identifies instance elements and produces a complete HD map.

\subsection{Ablation Study}
Aside from comparing against state-of-the-art methods, we further ablate different components of our DTCLMapper. 
Firstly, the core blocks (VPPSM, AIFCL, MCL, and PE) are explored separately. 
Then, the parameters in consistent learning are analyzed. Then, the basic framework and matching loss are analyzed.
We also visualize the preselected vector points for comparison. The map occupancy loss is further compared to other loss functions. Besides, other temporal fusion methods and the merging of temporal maps are discussed separately.

\begin{table}[t]
\fontsize{9}{12.8}\selectfont
\begin{center}
\caption{Results about different methods of the view transformer.}
\label{depth}
\begin{tabular*}{0.45\textwidth}{@{\extracolsep{\fill}}lcccc}
\toprule [2pt]   
Backbone    & $\rm{AP_{d}}$         & $\rm{AP_{p}}$         & $\rm{AP_{r}}$ & mAP  \\ \midrule [1pt]
GKT   & 53.9  & 51.2  & 55.8  & 53.6 \\
LSS &53.0 &46.4 &52.1 &50.5 \\
LSS+DepthGT & 54.9  & 51.7  & 57.9  & 54.8 \\ \bottomrule [2pt]
\end{tabular*}
\end{center}
\vspace {-1.6em}
\end{table}

\begin{figure}[tb]
      \centering
      \includegraphics[scale=0.4]{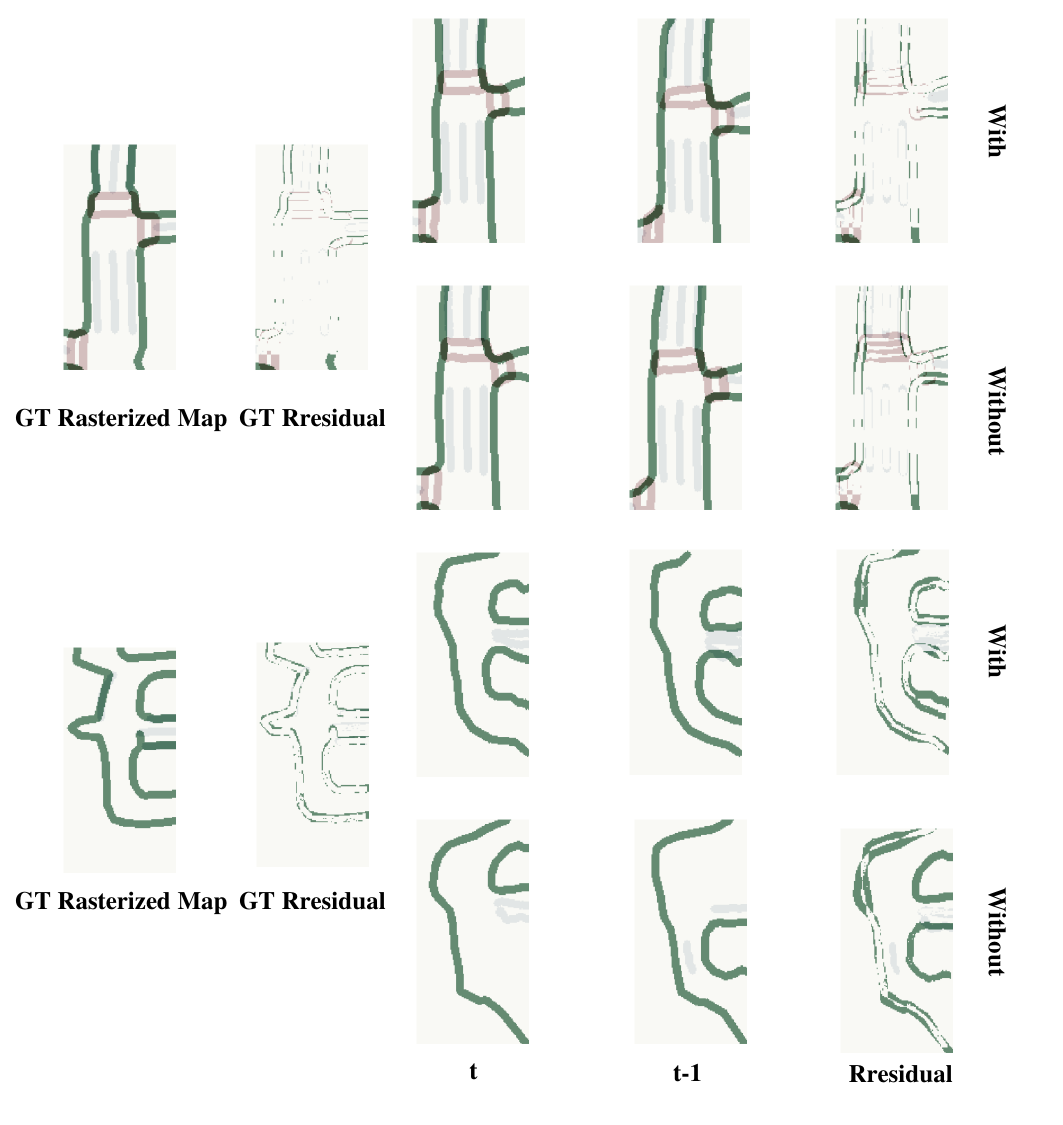}
      \vspace {-1.5em}
      \caption{Visual analysis of MO Loss. ``With'' means MO Loss is used, whereas ``Without'' means it is not used. Two comparison experiments in different scenarios are shown. Rasterized maps are derived through vectorized instances, while residual maps represent the difference between maps from consecutive frames.}
      \label{fig.moloss}
       \vspace {-1.5em}
\end{figure}
\textbf{Analysis of Core Blocks:}
Table~\ref{ablation} represents the experiment results of the core blocks. The baseline is a model based on transformers~\cite{maptr}. 
Subsequent experiments are realized on this basis. 
We start with the experiment with APPSM and AIFCL, as core modules in the ICL component. 
Each of them gives the model an accuracy boost (${+}1.3\%$ and ${+}0.9\%$). 
MCL component brings an improvement of ${+}0.4\%$ to the model. Finally, a geometric position encoder is added to VPPSM, which leads to the best result of $53.6\%$ in mAP. 
Meanwhile, we present the inference speed of each module. The proposed method only increases computation in the VPPSM and PE modules. From the results, our method shows comparable efficiency to the baseline.
In addition, we analyze the effect of the view transformer module. In contrast to transformer-based for 3D-to-2D BEV generation, LSS leverages perspective depth estimation in conjunction with intrinsic and extrinsic parameter projections to derive BEV features. Consequently, depth estimation plays a pivotal role in BEV generation within the LSS approach. As shown in Table~\ref{depth}, LSS without depth supervision performs less effectively than the transformer-based method, but LSS with depth supervision can promote the transformation between perspective and BEV view, demonstrating the guiding effect of depth supervision.

\begin{table}[t]
\caption{The ablation experiments on framework.}
\vspace {-1.0em}
\fontsize{9}{13.8}\selectfont
\label{tab:frame}
\begin{center}
\begin{tabular*}{0.45\textwidth}{@{\extracolsep{\fill}}l|ccccc}
\toprule [2pt] 
Epo   & 12   & 12   & 12   & 12   & 12   \\
Layer & 4    & 6    & 7    & 6    & 6    \\
Inst  & 50   & 50   & 50   & 40   & 30   \\ \hline
mAP   & 49.0 & 53.1 & 53.2 & 48.3 & 52.1 \\ \bottomrule [2pt]

\end{tabular*}
\end{center}

\caption{The ablation experiments on hybrid loss.}
\vspace {-1.0em}
\fontsize{9}{13.8}\selectfont
\label{tab:hloss}
\begin{center}
\begin{tabular*}{0.45\textwidth}{@{\extracolsep{\fill}}l|ccccc}
\toprule [2pt] 
Epo & 12   & 12   & 12   & 12   & 12   \\
T   & 120  & 240  & 300  & 250  & 200  \\
K   & 6    & 6    & 6    & 5    & 4    \\ \hline
mAP & 49.5 & 51.5 & 53.1 & 52.4 & 52.0 \\ \bottomrule [2pt]
\end{tabular*}
\end{center}

\fontsize{9}{12.8}\selectfont
\begin{center}
\caption{Results about the number of key embedding in IL.}
\label{vnum}
\begin{tabular*}{0.45\textwidth}{@{\extracolsep{\fill}}lcccc}
\toprule [2pt]  
The max number of `v' & $\rm{AP_{d}}$         & $\rm{AP_{p}}$         & $\rm{AP_{r}}$ & mAP  \\ \midrule [1pt] 
1                             & 63.8  & 55.9  & 59.1  & 59.6 \\
3                             & 64.7  & 56.3  & 60.1  & 60.4 \\
6                             &  63.5 &  56.9 &  59.6 &  60.0    \\ \bottomrule [2pt]
\end{tabular*}
\end{center}
\vspace {-2.0em}
\end{table}

\textbf{Analysis of the Number of Decoder Layer:}
In Table~\ref{tab:frame},  we tested the impact of different numbers of decoder layers on accuracy. While a $7$-layer decoder achieves the highest accuracy, it only improves by $0.1\%$ compared to a $6$-layer decoder and introduces a significant increase in parameter count. Therefore, we opt for a $6$-layer decoder structure. 

\textbf{Analysis of the Number of Instance Query:}
Under the constrained experimental conditions, we tested instance numbers of $30$ per frame, $40$ per frame, and $50$ per frame. Table~\ref{tab:frame} presents the experimental results, showing that the setting of $50$ instances per frame yields the highest gain.

\textbf{Analysis of One-to-Many Loss:}
We also conducted ablation experiments on two parameters, $K$ (indicating the number of query repetitions) and $T$ (representing the additional number of queries during training). In Table~\ref{tab:hloss}, it can be observed that $K{=}6$ and $T{=}300$ reach the best accuracy. 

\textbf{Analysis of Embedding:}
The number of embedding `$v$' plays an important role in contrastive learning. Conventionally, the larger the number of keys, the easier it is to learn implicit instance features. However, if there are many repeated and low-quality features, the efficiency of learning is geological. Therefore, we analyze the efficiency of learning by discussing the optimal number. We compared three sets of values, $1$, $3$, and $6$ for each label. Obviously, the accuracy improvement reaches its peak when the number reaches $3$. 
Ultimately, we choose the optimal number of `$v$' for each label to be $3$. 

\textbf{Visualization of PreSelected Vector Points:}
Fig.~\ref{fig.prechoose} presents a visualization of the preselected vector points. We select the reference point from the initial layer, the second layer, and the fourth layer of optimization to realize comparison. It can be found that compared with the random initialization setting, the initial vector points are near instances in the proposed VPPSM. 
Observing the distribution of the vector points in the subfigures at different layers, compared with another method, the vector points can regress to the correct instance more quickly and accurately under this initialization setting. 
Moreover, as shown in the subfigures of the predicted vectorized maps, the proposed method can obtain more accurate map instances.
\begin{table}[t]
\fontsize{9}{12.8}\selectfont
\begin{center}
\caption{Ablation results on Occupancy Loss.}
\label{ocloss}
\begin{tabular*}{0.45\textwidth}{@{\extracolsep{\fill}}lcccc}
\toprule [2pt] 
MO Loss   & Epoch & View-Trans & ICL & mAP  \\  \midrule [1pt] 
L1-Loss   & 24    & Transformer  & \checkmark     & 52.9 \\
Dice-Loss & 24    & Transformer  & \checkmark    & 51.9 \\
IoU-Loss  & 24    & Transformer  & \checkmark    & 52.1 \\ \bottomrule [2pt]
\end{tabular*}
\end{center}
\vspace {-1.0em}
\end{table}
\begin{table}[t]
\fontsize{9}{12.8}\selectfont
\begin{center}
\caption{Results about different temporal fusion methods.}
\label{temp}
\begin{tabular*}{0.45\textwidth}{@{\extracolsep{\fill}}lcccc}
\toprule [2pt]  
Method              & $\rm{AP_{d}}$         & $\rm{AP_{p}}$         & $\rm{AP_{r}}$ & mAP  \\ \midrule [1pt] 
Baseline            &62.7	&55.0	&58.5	&58.8 \\
Self-Attention      & 55.4  & 49.8  & 51.8  & 52.3      \\
GRU                 & 50.6  & 45.5  & 51.7  & 49.3      \\
Consistent Learning & 64.7  & 56.3  & 60.1  & 60.4 \\ \bottomrule [2pt]
\end{tabular*}
\end{center}
\vspace {-2.0em}
\end{table}
\begin{table}[t!]
\fontsize{9}{12.8}\selectfont
\begin{center}
\caption{Results about the efficiency of different methods. ``VT'' mans the translated methods between perspective and BEV views. ``Trans'' means Transformer construct. ``SA'' means self-attention.}
\label{effi}
\begin{tabular*}{0.45\textwidth}{@{\extracolsep{\fill}}lcccccc}
\toprule [2pt] 
                    & ~\cite{maptr}  & SA & GRU & ~\cite{maptrv2} & Our-b & Our-l   \\ \midrule [1pt] 
VT  & Trans & Trans   & Trans    & LSS     & Trans & LSS\\
mAP & 58.8  & 52.3    & 49.3     & 64.5    & 60.4  & 65.1   \\
FPS & 16.5  &  16.4    &  14.7     & 15.7     & 16.0   & 15.3     \\\bottomrule [2pt]   
\end{tabular*}
\end{center}
\vspace {-2.0em}
\end{table}

\textbf{Analysis of Map Occupancy Loss:}
The MO loss constrains the geometric position of all instances by their occupancy state globally, which is a self-supervision without labels. 
Three loss functions are compared to construct our optimal MO loss, as shown in Table~\ref{ocloss}. Here, the resolution of a grid map is $0.15m$, which is a default value.
It shows that L1-loss is the best choice, which can play the role of the MO loss precisely. 
Fig.~\ref{fig.moloss} further illustrates the visual analysis with or without using MO Loss, and the results further indicate the proposed MO Loss effectively narrows the gap between consecutive frames and produces higher-quality maps.

\textbf{Analysis of Temporal Fusion:}
The role of temporal information is a core topic in BEV research. 
In this part, we analyze the influence of mainstream temporal fusion methods on static map detection. Two methods are selected as research objects, \textit{i.e.}, self-attention~\cite{BEVFormer} and GRU~\cite{gru}. 
In the experiment, aligned features from consecutive frames are merged in the BEV feature layer first. 
Note that the first frame without historical information regards this frame information as history. 
Table~\ref{temp} shows the result that two temporal fusion methods produce a negative effect on detecting map elements. 
Upon examining the BEV features of different methods (as depicted in Fig.~\ref{Fig.intro}), it becomes evident that the features of temporal fusion methods are ambiguous around map instances, consequently leading to reduced efficiency in instance point regression. In contrast, our method exhibits distinct features surrounding instances.
This resonates with our findings concerning different methods in Fig.~\ref{Fig.1}. 
An indiscriminate fusion method leads to redundant features, particularly for static and similar features, thus impacting the accurate learning of essential elements.
However, the proposed consistent learning, concentrated on instances, can improve map detection. It can drive the perspective of studying temporal problems from the view of consistency.

\textbf{Analysis of Efficiency:} 
The update frequency is important in studying online vectorized map construction for autonomous driving. Table~\ref{effi} presents the analysis of the efficiency of different methods which are based on MapTR or MapTRv2. The transformer-based methods are more efficient compared with the LSS methods.  Additionally, our approach demonstrates comparable efficiency to state-of-the-art methods while achieving top accuracy. 

\begin{figure}[t!]
      \centering
      \includegraphics[scale=0.40]{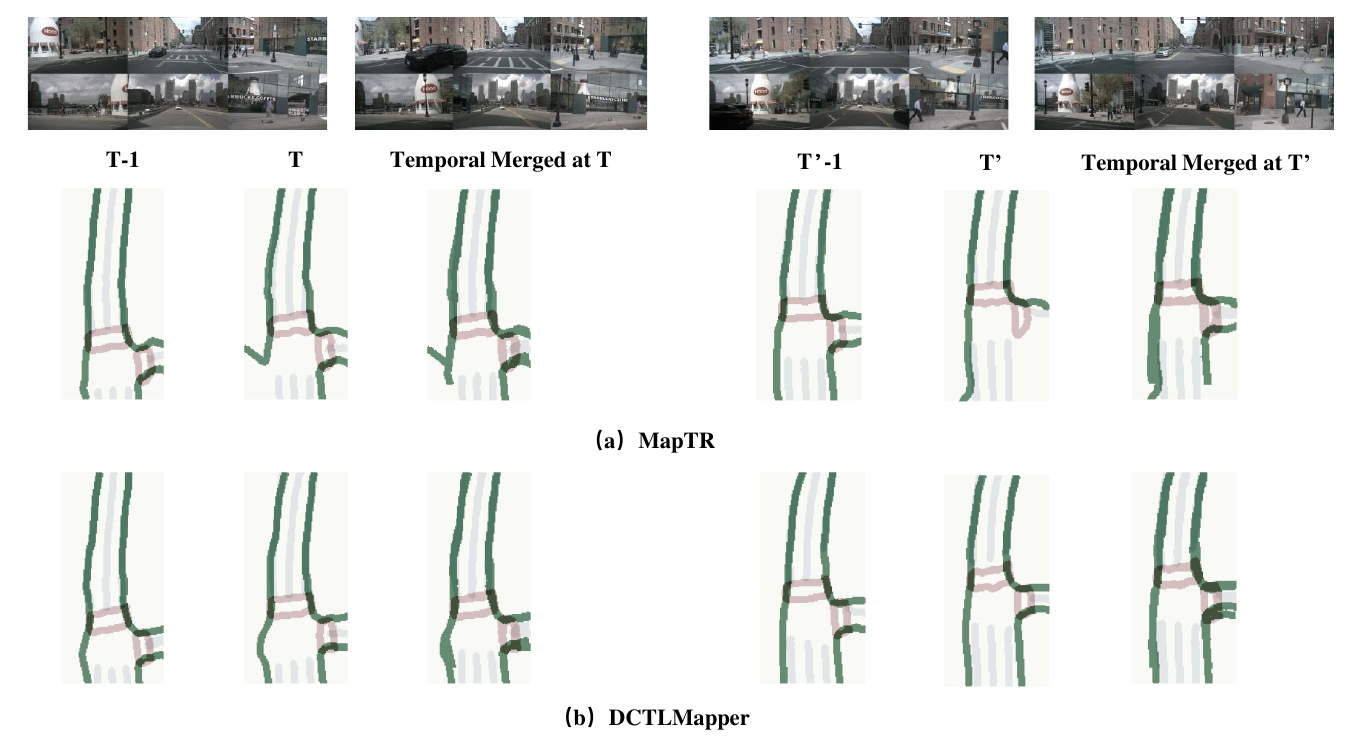}
      \vspace {-1.5em}
      \caption{Comparison of merging effects for HD grid maps. These grid maps are rasterized from corresponding vectorized maps. The merge is to add the grid map in each frame directly.}
      \label{fig.mergecmp}
      \vspace {-1.3em}
\end{figure}
\begin{figure}[tb]
      \centering
      \includegraphics[scale=0.30]{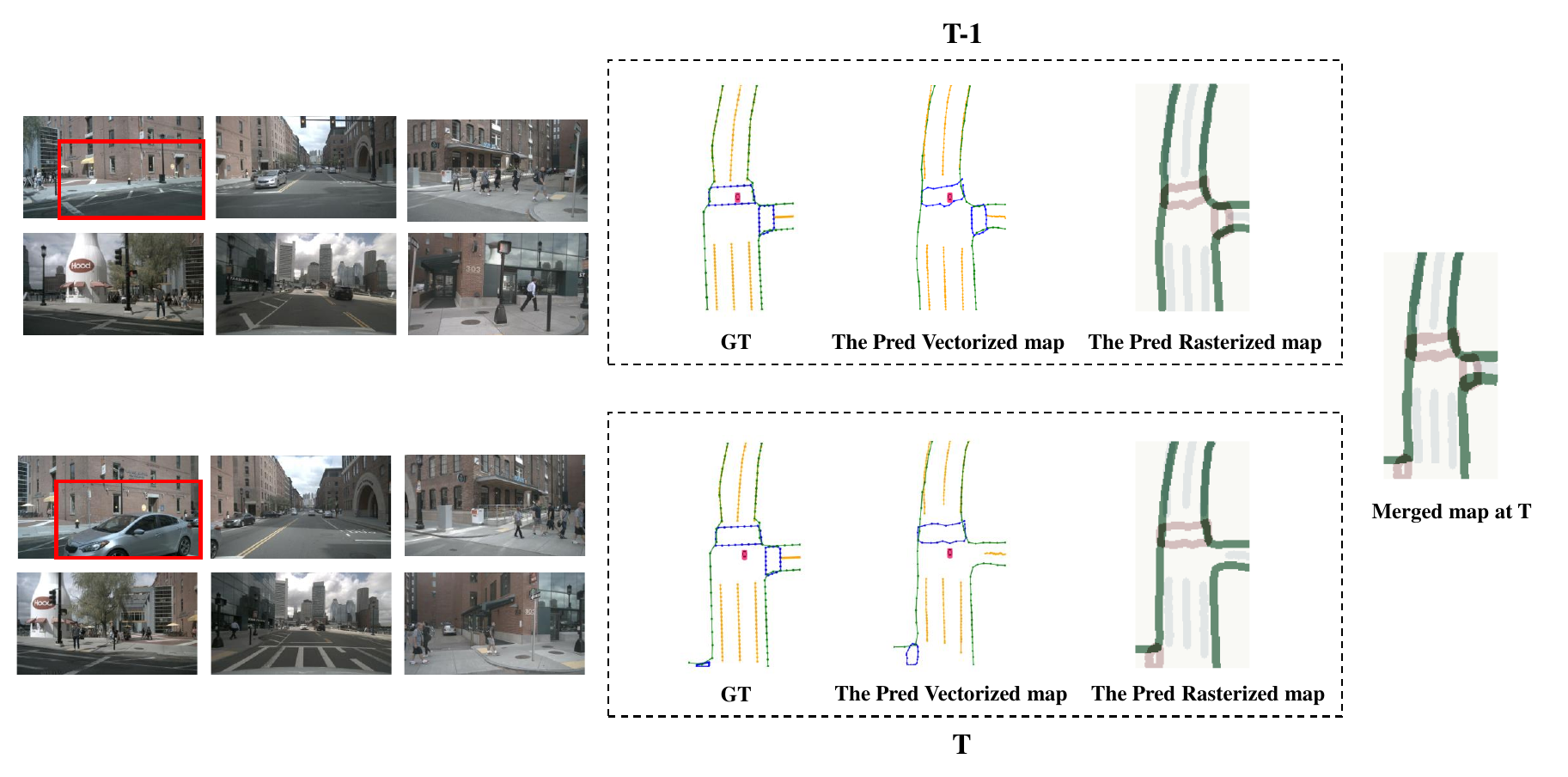}
      \vspace {-2.0em}
      \caption{Analysis of merging grid maps. When large object occlusions appear in perspective view, detecting the instance behind this object in a single frame is difficult. Temporal fusion of map layers can solve this difficulty.}
      \label{fig.mergemap}
     \vspace {-1.5em}
\end{figure}
\textbf{Discussion of Merging Temporal Maps:}
While the proposed temporal consistency method significantly enhances map generation accuracy, we delve further into temporal fusion methods for map layers.
In an HD map, the absolute location of each map element is fixed in a global environment, meaning that local HD maps of successive moments can be merged simply through ego-motion transformations. 
Thus, we explore whether the temporal merging of maps can alleviate the sparsity caused by object occlusions.  
Fig.~\ref{fig.mergecmp} depicts a set of the merging results of local HD maps. 
It can be found that incorrect instance detection in a single frame can lead to map fusion. Our method improves the accuracy of single-frame map detection, producing more accurate maps after fusion.
Although our method has enhanced the representation of weak instances to a large extent, there are extreme cases where detecting instances is greatly difficult, such as a large vehicle blocking key feature learning in the perspective view as shown in Fig.~\ref{fig.mergemap}. 
Encountering this situation, merging temporal maps can obtain accurate map elements from the previous local map. 
This method of temporal fusion for map layers is not strictly tied to the network architecture, which can effectively enhance any vectorized HD map construction model. 
As shown in Fig.~\ref{fig.mergecmp}, MapTR replenishes the map instances that are missing in the current perception through the temporal fusion of map layers.
These results underscore the effectiveness of local map consistency learning while emphasizing the completed expression achieved through local map temporal fusion.

\section{Conclusion}
Considering that existing temporal fusion methods are not friendly to map instances, this paper proposes a novel solution for consistent learning of temporal instances in vectorized HD map construction.
We devise the DTCLMapper framework to realize consistent learning for detecting map instances, combining the instance consistent learning and the map consistent learning. 
The two consistency learning components are studied progressively, the former faces semantic learning of local instances, and the latter learns the geometric distribution of global instances. Moreover, the instance consistent learning aggregates the fast regression design of vector points and the temporal contrastive learning of aggregated instance features, ensuring high accuracy and temporal richness of instances.
The proposed method achieves state-of-the-art performance in constructing vectorized HD maps, reaching $61.9\%$ and $65.1\%$ mAP scores on the nuScenes and Argoverse datasets, respectively. 
Simultaneously, this paper discusses the effectiveness of merging temporal maps. Compared with temporal fusion in the feature layer, merging maps can fuse the missing instances more intuitively and improve the quality of maps.

There is still rich research space to further enhance HD map construction in real-world driving scenarios.
The presented local map merging is conducted by a simple fusion operation and a ground truth ego motion is limited to the current frame.
In the future, we will explore how to merge multi-frame maps according to the probability model to realize the generation of a wider range of maps and optimize ego-motion estimation at the same time, similar to the research of simultaneous localization and mapping.

\bibliographystyle{IEEEtran}
\bibliography{refer.bib}

\end{document}